\documentclass[year,preprint,review,12pt]{elsarticle}

\usepackage{amssymb}
\usepackage{amsmath}
\usepackage{lineno}

\usepackage{hyperref}

\usepackage{tikz}
\usepackage{pgfplots}
\pgfplotsset{compat=1.18}
\usepackage{xcolor,colortbl}
\pgfplotsset{compat=newest}
\usepackage{float} 
\usepackage{longtable}
\usepackage{multirow}
\usepackage{booktabs}
\usepackage{comment}

\usepackage[most]{tcolorbox}
\tcbuselibrary{listingsutf8}
\usepackage{listings}

\usepackage{amsmath}

\usepackage{latexsym}

\usepackage[T1]{fontenc}
\usepackage[utf8]{inputenc}

\usepackage{microtype}

\usepackage{inconsolata}

\usepackage{graphicx}

\usepackage{amssymb}% http://ctan.org/pkg/amssymb
\usepackage{pifont}% http://ctan.org/pkg/pifont.bib

\definecolor{onesteporange}{RGB}{244,148,52}     % orange!90 approx
\definecolor{structuredgreen}{RGB}{73,190,48}      % green!70!black approx
\definecolor{distilledred}{RGB}{249,62,57}       % red!90 approx

\newcommand{\Sionestep}{$S_i^\text{one-step}$}
\newcommand{\Sidistill}{$S_i^\text{distill}$}
\newcommand{\Sistruct}{$S_i^\text{struct}$}

\newcommand{\cmark}{\ding{51}}%
\newcommand{\xmark}{\ding{55}}%
\definecolor{niceorange}{HTML}{F86624}

\journal{JAMIA Open}

\begin{document}
\begin{frontmatter}

%% Title
\title{\texttt{DistillNote}: Toward a Functional Evaluation Framework of LLM-Generated Clinical Note Summaries}

%% Authors and affiliations
\author[1,2]{Heloisa Oss Boll, MSc.}
\author[3]{Antonio Oss Boll, BSc.}
\author[3]{Leticia Puttlitz Boll, BSc.}
\author[1,2]{Ameen Abu-Hanna, PhD.}
\author[1,2]{Iacer Calixto, PhD\corref{cor1}}
\cortext[cor1]{Corresponding author}
\ead{i.coimbra@amsterdamumc.nl}

\affiliation[1]{organization={Department of Medical Informatics, Amsterdam UMC, University of Amsterdam},
            %addressline={}, 
            city={Amsterdam},
            country={The Netherlands}}

\affiliation[2]{organization={Amsterdam Public Health, Methodology},
            %addressline={},
            city={Amsterdam},
            country={The Netherlands}}

\affiliation[3]{organization={Institute of Mathematics and Statistics, University of São Paulo},
            %addressline={},
            city={São Paulo},
            country={Brazil}}

%% Abstract
\begin{abstract}
% structure into Objective, Methods, Results, Conclusion; up to 300 words
\textbf{Objective:} Large language models (LLMs) are increasingly used to generate summaries from clinical notes. However, their ability to preserve essential diagnostic information remains underexplored, which could lead to serious risks for patient care. This study introduces \texttt{DistillNote},\footnote{We will publicly release all code and datasets upon acceptance.} an evaluation framework for LLM-generated summaries that targets their functional utility by applying the generated summary downstream in a complex clinical prediction task, explicitly quantifying how much prediction signal is retained.

\textbf{Methods:} We generated over 192,000 LLM summaries from MIMIC-IV clinical notes with increasing compression rates: standard (36\%), section-wise (53\%), and distilled section-wise (79\%). Heart failure diagnosis was chosen as the downstream prediction task, as it requires integrating a wide range of clinical signals. State-of-the-art LLMs were fine-tuned on both the original notes and their summaries, and their diagnostic performance was compared using the Area Under the Receiver Operating Characteristic Curve (AUROC). We contrasted \texttt{DistillNote}’s results with evaluations from LLM-as-judge and clinicians, assessing consistency across different evaluation methods.

\textbf{Results:} Summaries generated by LLMs maintained a strong level of heart failure diagnostic signal despite substantial compression. Models trained on the most condensed summaries (about 20 times smaller) achieved an AUROC of 0.92, compared to 0.94 with the original note baseline (97\% retention). Functional evaluation provided a new lens for medical summary assessment, emphasizing clinical utility as a key dimension of quality.

\textbf{Conclusion:} \texttt{DistillNote}  introduces a new scalable, task-based method for assessing the functional utility of LLM-generated clinical summaries. Our results detail compression-to-performance tradeoffs from LLM clinical summarization for the first time. The framework is designed to be adaptable to other prediction tasks and clinical domains, aiding data-driven decisions about deploying LLM summarizers in real-world healthcare settings.

Word count: 3998
\end{abstract}



%% Keywords

\end{frontmatter}

%\begin{linenumbers}

\section{Lay summary}
%A requirement from JAMIA Open. Up to 200 words.
Healthcare workers produce large numbers of written records about patients. To manage this information, large language models (LLMs) are increasingly used to create summaries of lengthy patient notes. However, a critical question remains: to what extent do these LLM summaries retain the essential details required for important clinical predictions? And at what cost?

This study introduces \texttt{DistillNote}, a framework to test whether LLM-generated summaries retain critical information. Our hypothesis is: if LLM summaries keep the key medical details, models trained on these summaries should predict clinical outcomes almost as well as when using the full notes. This approach is useful because it 1) connects evaluation to real clinical tasks and 2) can be scaled to thousands of summaries.

Using real hospital records, we generated summaries at different lengths to test compression tradeoffs. Heart failure diagnosis served as our case study as it requires integrating multiple clinical signals. We found that even the most compressed summaries (20x shorter) retained 97\% of the diagnostic performance of full notes, detailing useful tradeoffs.

\texttt{DistillNote} provides an objective way to evaluate LLM summaries before deployment. By focusing on whether summaries support real clinical tasks, it helps ensure patient safety while improving documentation efficiency.


\section{Introduction}
\label{intro}
Clinical notes detail a patient's journey in the hospital, being an important part of Electronic Health Records (EHR). However, notes are often lengthy, redundant, and difficult to navigate under time constraints. Clinicians may find dozens of pages of clinical text when making time-critical decisions, causing reduced quality of care and cognitive overload \cite{nijor2022patient, tajirian2020influence}. Moreover, over half of a clinical note's content may be duplicated \cite{steinkamp2022prevalence}, and nearly 75\% of clinicians identify EHR as a major burnout cause \cite{tajirian2020influence}.

Summarization with large language models (LLMs) offers a potential remedy, with recent work showing that LLMs match or surpass human experts in clinical summarization \cite{VanVeen:23, schoonbeek_completeness_2024}. Yet there is no standard framework for systematically evaluating clinical AI summaries \cite{subramanian2025clinical, gebauer_benchmarking_2025}. Common automatic evaluation approaches include $n$-gram-based and embedding-based metrics \cite{vanSchaikPugh24}, which focus on lexical and semantic similarity, respectively, but overlook clinical quality aspects. Human expert assessment can provide high-quality judgments but is time-consuming, costly, and difficult to scale \cite{croxford_current_2025}. More recently, LLMs are used directly to evaluate summaries (i.e., often referred to as LLM-as-judge or autograders \cite{bavaresco-etal-2025-llms}), which can offer scalability but suffer from inconsistency, bias, and limited reliability across domains \cite{schroeder_can_2024}.

In-the-wild generative AI evaluation should also assess model impact on downstream patient outcomes \cite{Jabbour2025EvaluationFF}, however scalable, systematic methods targeting real clinical outcomes remain unexplored. This evaluation gap becomes especially pressing as EHR vendors begin deploying LLM summaries in real world, where thousands of outputs are used by doctors directly within patient care \cite{epic2025ai, epic2023generative, silberlust2025ai}. Without frameworks tailored for evaluating clinically relevant information retention at scale, there is a high risk of deploying LLM summarizers that appear coherent but fail to preserve essential information.

%hypotheses focus -- llm focused prompts increase focus. why did you do things?

To address this gap, we introduce \texttt{DistillNote}, a conceptual framework for evaluating LLM clinical summaries through the lens of \textit{clinical signal preservation} by directly measuring how compressed notes maintain their utility for clinical prediction tasks. We investigate two fundamental questions: (1) \textit{Do LLM summaries preserve sufficient diagnostic information to maintain prediction performance on a complex clinical task?} and (2) \textit{How do compression-utility trade-offs vary across models and compression rates?}

Towards this purpose, we generate $>$192k summaries from MIMIC-IV clinical notes with multiple state-of-the-art LLMs, comparing outputs with different compression rates (36-53-79\%). Concretely, we investigate the feasibility of our conceptual framework using heart failure (HF) as the downstream clinical task, as HF represents a complex diagnostic challenge requiring integration of multiple clinical signals, and which early identification is essential for effective disease management and burden reduction \cite{chen_exploring_2024}. Results demonstrate that summaries preserve 93-99\% of original diagnostic performance (i.e., AUROC), indicating that LLMs \textit{can preserve complex diagnostic signals} even at substantial compression rates, albeit with tradeoffs that must be carefully considered before deployment.

Our main contributions include: (1) A novel conceptual framework to evaluate LLM clinical summaries at scale through the lens of diagnostic signal retention; (2) An implementation of this framework whereby we use HF as the downstream clinical task, including a comparison of findings with LLM-as-judge and clinician evaluation across three metrics; (3) A publicly available dataset of $>$192k LLM-generated summaries to support clinical NLP research.

\begin{figure}[t!]
    \centering
    \includegraphics[width=\textwidth]{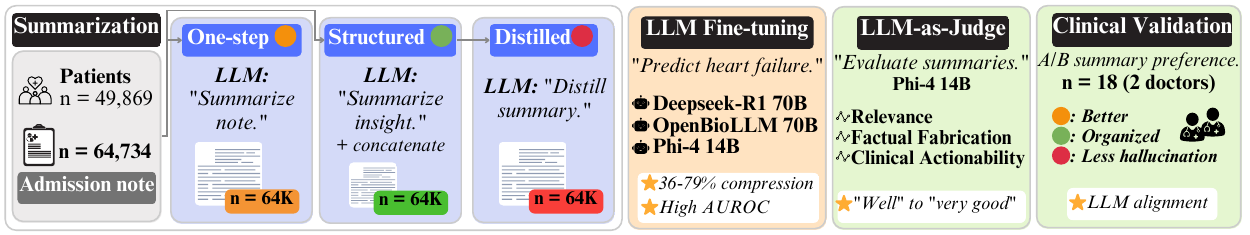}
    \caption{\textbf{Overview of \texttt{Distillnote}}. We introduce evaluating clinical LLM summary quality at scale based on functional utility, i.e. using summaries to diagnose heart failure, comparing the outputs with an LLM judge and clinician evaluations. We observe retention of most predictive performance with \textit{substantial text compression}, indicating LLM summaries preserve \textit{critical diagnostic signals}.}
    \label{fig:distillnote}
\end{figure}

\section{Background and Significance}
Recent studies show that large language models (LLMs) can generate clinical summaries that match or outperform human experts in completeness, conciseness, and correctness \cite{VanVeen:23, schoonbeek_completeness_2024}. However, these do not assess whether the summaries retain the diagnostic signals necessary for downstream clinical predictions.

Conversational and section-specific summarization has broadened the scope of clinical summarization \cite{li2024improving, chen_exploring_2024, liang_novel_2019}, but still lacks objective measures of how well summaries preserve clinically relevant information. Efforts to improve faithfulness and reduce hallucinations through calibration and data-centric training \cite{hegselmann2024annotations, hegselmann2024datacentric} increase summary quality but offer little insight into clinical usefulness.

Although LLM-generated summaries have shown potential for supporting downstream applications such as diagnostic reasoning and risk prediction \cite{choudhuri2025summarizing, Verma2025VerifiableSO}, no systematic evaluation has linked summary quality to functional clinical performance, nor evaluated compression-to-performance tradeoffs at scale.  

\texttt{DistillNote} addresses these gaps by introducing a generalizable and scalable framework to quantify diagnostic signal preservation in LLM-generated summaries, enabling more trustworthy and clinically applicable use of LLM summarizers.

\section{Materials and Methods}
Let $\{A_i\}_\text{i=1}^M$
denote the set of $M$ \textit{admission notes}, where each is represented as \( A_i = \{t_1, t_2, \cdots, t_N\} \), \( t_j \) is the \( j \)-th token  and \( N \) is the total number of tokens in note $A_i$.

\subsection{Dataset} 

Our prediction task requires clinical notes written at hospital admission.\footnote{We use admission notes as they represent the earliest clinical documentation in the patient visit. This enables anticipating complications during the current hospitalization.} To avoid leakage from future sections, we truncate MIMIC-IV discharge summaries ~\cite{Johnson2023,johnson_mimic-iv-note_nodate} into \textit{admission notes} as per \cite{ rohr_revisiting_2024}, retaining \textit{chief complaint}, \textit{history of present illness}, \textit{medical history}, \textit{admission medications}, \textit{allergies}, \textit{physical exam}, \textit{family history}, and \textit{social history} sections.
Our final dataset consists of 64,734 notes from 49,869 patients (more details in \S~\ref{appendix:admnotestats}).

\subsection{Large language models (LLMs)}
\label{sec:llms}

We experiment with three LLMs in our framework:
%to generate summaries (\S \ref{sec:summarization}) and to predict HF (\S \ref{sec:hf_prediction}):
\texttt{Deepseek-R1-70B}\footnote{\url{https://huggingface.co/Deepseek-ai/Deepseek-R1-Distill-Llama-70B}}  (reasoning) ,\texttt{OpenBioLLM-70B}\footnote{\url{https://huggingface.co/aaditya/Llama3-OpenBioLLM-70B}} (biomedical), 
and \texttt{Phi-4-14B}\footnote{\url{https://huggingface.co/microsoft/phi-4}} (general).
Each has shown strong performance in complex reasoning tasks and improved alignment with human preferences \cite{abdin_phi-4_2024, lee_distillation_2025, subramanian_small_2025, OpenBioLLM, grattafiori_llama3_2024, Deepseek_r1_2025}.

\subsection{LLM-based summarization}\label{sec:summarization}
We use all LLMs in \S~\ref{sec:llms} to generate clinical note summaries.

\paragraph{One-step summarization}
A \textbf{one-step summary} \Sionestep is produced by generating the summary of an admission note \( A_i \) in one shot:
\begin{equation}
    \mathcal{S}_i^\text{one-step} = \text{LLM}([q_\text{one-step}; A_i]), \label{eq:onestep-summarisation}
\end{equation}  
where \( q_\text{one-step} \) is a summarization prompt, and \([]\) denotes concatenation.

\paragraph{Divide-and-conquer summarization}

We define a set of four prompts
$\{P_k\}_\text{k=1}^{4}$,
each targeting one clinical insight at admission: \textit{chief complaint}, \textit{medical history}, \textit{exam findings}, and \textit{social/family background}.
For each note \( A_i \), the prompts
$\{P_k\}_\text{k=1}^{4}$
are processed independently by an LLM:  
\begin{equation}
    t_\text{i,k} = \text{LLM}([P_k; A_i]),
\end{equation}  
where \( t_\text{i,k} \) represents the LLM-generated summary for clinical insight \( k \) given an admission note $A_i$.
We generate a \textbf{structured summary} \( \mathcal{S}^{\text{struct}}_i \) by concatenating all four intermediate summaries for note $A_i$ as follows:\footnote{We add section headers for each clinical insight $k$ in \( \mathcal{S}^{\text{struct}}_i \), but omit these in Eq.~(\ref{eq:structured-summarisation}) to avoid clutter.}
\begin{equation}\label{eq:structured-summarisation}
    \mathcal{S}^{\text{struct}}_i = [s_\text{i,1}; s_\text{i,2}; s_\text{i,3}; s_\text{i,4}].
\end{equation}  

\begin{comment}
Next, to produce the final summary \( S^{\text{distill}}_i \) for note $A_i$, a final summarization prompt \( g \) is defined as:
\begin{equation}
    g = [q_g; \mathcal{S}_i^{\text{struct}}],
\end{equation}  
where \( q_g \) is the final summarization query. The final summary \( S_{\text{conquer}} \), reflecting the patient's clinical trajectory, is generated as:  
\begin{equation}
    S_{\text{conquer}} = \text{LLM}(g).
\end{equation}
\end{comment}
Next, we produce a \textbf{distilled summary} \( S^{\text{distill}}_i \) for note $A_i$ by using the structured summaries $\mathcal{S}^{\text{struct}}_i$ generated in the previous step as input:
\begin{equation}
S^{\text{distill}}_i = LLM([q_\text{distill}; \mathcal{S}_i^{\text{struct}}]),
\end{equation}
where \( q_\text{distill} \) is the distilled summarization query.

\begin{comment}
\paragraph{Overview}
We compare the three sets of summaries previously detailed for each note $A_i$.
\Sionestep{} is a standard summary generated with an LLM in one pass;
\Sistruct{} is a structured summary where clinical insights are summarized independently and then concatenated, and \Sidistill{} is the distilled summary that integrates the structured summaries into a single, final summary.
We provide details on the prompts used and sampling parameters in \ref{prompts} and \ref{samplingparams}, respectively. 
\end{comment}

\paragraph{Details}
Refer to \ref{prompts} for details on the prompts and \ref{samplingparams} for details on our hyperparameters.

%\subsection{Dataset}

%\begin{table}[H]
%\centering
%\begin{tabular}{lr}
%\toprule
%\textbf{Metric} & \textbf{Value} \\
%\midrule
%\# notes & 64,734 \\
%\# patients & 49,869 \\
%\bottomrule
%\end{tabular}
%\caption{Admission note cohort statistics.}
%\label{tab:icu_notes_summary}
%\end{table}

%We use discharge summaries from the MIMIC-IV dataset~\cite{Johnson2023,johnson_mimic-iv-note_nodate}, and parse and preprocess them into \textit{admission notes} as per \cite{ kweon_ehrnoteqa_2024, rohr_revisiting_2024}. Our final dataset consists of 64,734 admission notes from 49,869 patients (see \ref{appendix:admnotestats} for more details).

\subsection{Summary evaluation}
\paragraph{LLM-as-judge}
We implement an LLM-as-judge evaluation framework with \texttt{Phi-4}, due to its improved alignment with human preferences \cite{abdin_phi-4_2024, bavaresco_llms_2024}. We implement continuous scoring and take inspiration from G-Eval \cite{liu2023geval}, using the probability the LLM assigns to its scores to account for uncertainty (see \S~\ref{appendix:llm-judge} for details).

To assess summaries, we use three 1--5 scale metrics: \textbf{Relevance}, \textbf{Factual Fabrication} (both adapted from \cite{krolik_towards_2024}), and \textbf{Clinical Actionability}, which we propose in this work. Ranges go from ``very poor'' to ``perfect''. \textit{Relevance} measures coverage of essential medical details,
\textit{Fabrication} checks for hallucinations, and
\textit{Clinical Actionability} assesses support for medical decision-making.
LLM judge prompts for each metric include the original note, generated summary, metric definition, and few-shot examples (see \S~\ref{relevance}, \S~\ref{fabrication}, \S~\ref{actionability}). We perform ANOVA,  Tukey's HSD tests, and Cohen's d calculations to analyze result differences. % (\autoref{tab:geval-scores}).

\paragraph{Clinician validation}

Two board-certified clinicians participated in a blinded pairwise comparison study, evaluating 18 admission notes and their summaries. Clinicians were presented with the original note followed by two unlabeled summaries A and B (either \Sionestep, \Sistruct, or \Sidistill, presented in randomized order). For each pair, clinicians indicated their preference across the same three metrics as the LLM judge. Binomial tests were used to determine if preferences across cases were statistically significant with Bonferroni correction for multiple comparisons. A correlation analysis was performed to evaluate the link between LLM judge and human preferences.

\subsection{Heart failure (HF) prediction}
\label{sec:hf_prediction}

We fine-tune each LLM in \S~\ref{sec:llms} for HF prediction using as input the patient's full original admission note $A_i$, or one of the three summaries \Sionestep, \Sistruct, \Sidistill. The output is binary: 0 (negative) or 1 (positive) for the condition. Patients appear in only one split (train/val/test) and we use class stratification, i.e., 78\% negative vs. 22\% positive (see \S~\ref{finetuning} for details).
%Each model was trained separately on four input types: $A_i$ (original admission notes), \Sionestep, \Sistruct, and \Sidistill. 
Given the imbalanced nature of the task, AUROC and AUPRC are our main metrics \cite{NEURIPS2024_4df3510a}, whereas we also report F1-scores for completeness.

\section{Results}
\subsection{LLM-as-judge summary evaluation}

\begin{figure}[t]
\centering
\resizebox{0.7\textwidth}{!}{
\begin{tikzpicture}[font=\scriptsize\bfseries]
\begin{axis}[
    ybar,
    bar width=12pt,
    ymin=1, ymax=5.2,
    ylabel={LLM-as-Judge Score},
    ytick={1,2,3,4,5},
    yticklabels={
        1.0 (Very poor),
        2.0 (Poor),
        3.0 (Adequate),
        4.0 (Very good),
        5.0 (Perfect)
    },
    symbolic x coords={Factual\\Fabrication, Relevance, Clinical\\Actionability},
    xtick=data,
    xtick style={draw=none},
    x tick label style={font=\scriptsize\bfseries, align=center},
    tick label style={font=\scriptsize\bfseries},
    label style={font=\scriptsize\bfseries},
    grid=both,
    major grid style={line width=.2pt, draw=gray!30},
    minor grid style={line width=.1pt, draw=gray!10},
    axis lines=left,
    tick style={color=black!70},
    enlarge x limits=0.2,
    legend style={
        at={(0.5, -0.27)},
        anchor=south,
        legend columns=3,
        font=\scriptsize\bfseries,
        draw=black,
        fill=white
    }
]

% One-step: orange
\addplot+[
    fill=onesteporange,
    draw=black,
    line width=0.4pt,
    error bars/.cd,
        y dir=plus,
        y explicit
] coordinates {
    (Factual\\Fabrication, 3.75) +- (0, 0.28)
    (Relevance,  4.19) +- (0, 0.15)
    (Clinical\\Actionability, 3.85) +- (0, 0.22)
};

% Structured: green
\addplot+[
    fill=structuredgreen,
    draw=black,
    line width=0.4pt,
    error bars/.cd,
        y dir=plus,
        y explicit
] coordinates {
    (Factual\\Fabrication, 3.70) +- (0, 0.31)
    (Relevance,  3.96) +- (0, 0.18)
    (Clinical\\Actionability, 3.53) +- (0, 0.24)
};

% Distilled: red
\addplot+[
    fill=distilledred,
    draw=black,
    line width=0.4pt,
    error bars/.cd,
        y dir=plus,
        y explicit
] coordinates {
    (Factual\\Fabrication, 3.92) +- (0, 0.26)
    (Relevance,  3.99) +- (0, 0.20)
    (Clinical\\Actionability, 3.53) +- (0, 0.28)
};

\legend{One-step, Structured, Distilled}
\end{axis}
\end{tikzpicture}
}
\caption{\textbf{LLM-as-judge scores across summarization approaches.} 
All fall in the ``adequate'' to ``very good'' range. One-step scores highest on relevance and actionability, while Distilled shows higher factuality. Standard deviations indicate scoring uncertainty.}
\label{fig:scores}
\end{figure}

All summarization approaches achieved ``adequate'' to ``very good'' score ranges across all metrics (Fig.~\ref{fig:scores}).
Statistical analysis indicated significant differences between them (p~$<$~0.01, see \S~\ref{stats-llmjudge} for details): the One-step approach achieved the highest overall performance (3.93 $\pm$ 0.29) with strong relevance (4.19 $\pm$ 0.15) and clinical actionability (3.85 $\pm$ 0.22); the Distilled strategy demonstrated best factual reliability (3.92 $\pm$ 0.26) with medium-to-large effect sizes compared to One-step ($d$ = $0.62$) and Structured ($d$ = $0.75$); the Structured approach achieved slightly lower overall scores (3.73 $\pm$ 0.30) but had a more balanced performance.

\subsection{Clinician summary evaluation}

Clinicians' preferences show positive Spearman correlation with LLM judge evaluations ($\rho$ = 0.67, p < 0.05, \autoref{tab:llm_clinician}). Pairwise comparisons between clinicians were not statistically significant (p > 0.05), indicating alignment in preference patterns.  Clinicians preferred \Sionestep{} overall, and chose \Sionestep{} over \Sistruct{} 8 (vs. 7) out of 12 times for relevance (vs. actionability), and preferred \Sionestep{} over \Sidistill{} 9 out of 12 times for both metrics.  

A qualitative analysis of clinicians' comments pointed that \Sionestep{} were \textit{``relevant,''} \Sistruct{} \textit{``organized,''} and \Sidistill{} \textit{``short''} summaries that \textit{``suffice''} for clear cases, indicating different summaries exhibit different patterns of support for decision making. 

\subsection{Functional evaluation}

\begin{table}[t!]
\centering
\resizebox{0.6\textwidth}{!}{%
\begin{tabular}{lllccc}
\toprule
\textbf{Strategy} & \textbf{S?} & \textbf{Model} & \textbf{AUROC} & \textbf{AUPRC} & \textbf{F1} \\
\midrule
\multirow{3}{*}{Full note}
& \xmark & DS & 0.935 & 0.842 & 0.766 \\
& \xmark & OB & 0.939 & \textbf{0.852} & \textbf{0.771} \\
& \xmark & Phi-4 & \textbf{0.941} & \textbf{0.852} & \textbf{0.771} \\
\midrule
\multirow{3}{*}{One-step}
& \cmark & DS & 0.926 & 0.818 & \textbf{\underline{0.747}} \\
& \cmark & OB & 0.926 & 0.819 & 0.734 \\
& \cmark & Phi-4 & \textbf{\underline{0.929}} & \textbf{\underline{0.822}} & 0.745 \\
\cmidrule{2-6}
\multirow{3}{*}{Structured} 
& \cmark & DS & 0.922 & 0.804 & \textbf{0.739} \\
& \cmark & OB & 0.916 & 0.802 & 0.726 \\
& \cmark & Phi-4 & \textbf{0.926} & \textbf{0.808} & 0.730 \\
\cmidrule{2-6}
\multirow{3}{*}{Distilled} 
& \cmark & DS & \textbf{0.917} & 0.794 & \textbf{0.726} \\
& \cmark & OB & 0.911 & \textbf{0.799} & \textbf{0.726} \\
& \cmark & Phi-4 & 0.873 & 0.757 & 0.701 \\
\bottomrule
\end{tabular}}
\caption{
\textbf{Functional evaluation: heart failure prediction performance.}  
Summaries mostly preserve predictive performance, with AUROC > 0.9 and AUPRC > 0.8 (except for Distilled where results are slightly lower). While metrics are slightly reduced compared to full notes, diagnostic signal is largely retained.
\textbf{Bold}: best within each strategy.  
\underline{Underline}: best among summarization strategies.  
\textbf{S?}: Is a summary used as input?  
\textbf{DS}: DeepSeek-R1.  
\textbf{OB}: OpenBioLLM.
}
\label{tab:preds}
\end{table}

\begin{figure}[t!]
\centering
\begin{tikzpicture}[font=\tiny]
\begin{axis}[
    xlabel={Text compression (\%)},
    ylabel={AUROC},
    xmin=0, xmax=85,
    ymin=0, ymax=1,
    grid=both,
    width=0.65\columnwidth, 
    height=6cm,              
    font=\scriptsize\bfseries,
    xtick={0,20,40,60,80},
    ytick={0,0.2,0.4,0.6,0.8,1.0},
    major grid style={line width=.2pt,draw=gray!30},
    minor grid style={line width=.1pt,draw=gray!10},
    axis lines=left,
    tick style={color=black!70},
    every axis label={font=\scriptsize},
    title style={font=\scriptsize\bfseries, yshift=-0.1cm},
]
\addplot[only marks, mark=*, mark size=3pt, color=blue!80, line width=0.4pt] coordinates {(0, 0.939)};
\node at (axis cs:3,0.88) {\textbf{F}};
\addplot[only marks, mark=square*, mark size=3pt, color=onesteporange, line width=0.4pt] coordinates {(36.4, 0.927)};
\node at (axis cs:36.4,0.86) {\textbf{O}};
\addplot[only marks, mark=triangle*, mark size=3.2pt, color=structuredgreen, line width=0.4pt] coordinates {(52.8, 0.921)};
\node at (axis cs:52.8,0.86) {\textbf{S}};
\addplot[only marks, mark=diamond*, mark size=3.5pt, color=distilledred, line width=0.4pt] coordinates {(79.0, 0.900)};
\node at (axis cs:79.0,0.83) {\textbf{D}};
\end{axis}
\end{tikzpicture}
\caption{\textbf{Minimal loss in AUROC despite compression.}
Summarization strategies yield AUROC scores within 1.2--4.0\% of full-note baseline, even at 79\% text reduction. \textbf{F} = Full note (0.939 AUROC, 412 average words), \textbf{O} = One-step (0.927, 262 average words), \textbf{S} = Structured (0.921, 195 average words), \textbf{D} = Distilled (0.900, 87 average words).}
\label{fig:tradeoff}
\end{figure}
All summarization methods achieve high AUROC scores in heart failure prediction when compared to full notes, losing 1.2--4.0\% in performance despite 36--79\% compression. This indicates that they retain the most critical signals to distinguish high- from low-risk patients (\autoref{tab:preds}, \autoref{fig:tradeoff}). Efficiencies (compression-to-loss ratios) are high for all metrics (\autoref{tab:efficiency}). AUPRC and F1 show sharper tradeoffs, especially under higher compression (\autoref{tab:tradeoff}). Phi-4 performs best in the \Sionestep (0.929 AUROC) and \Sistruct (0.926) settings, while DeepSeek leads in the \Sidistill approach (0.917).  %(\autoref{tab:preds}, \autoref{fig:tradeoff}).

\section{Discussion}

\paragraph{Despite substantial compression, LLM summaries retain most HF diagnostic signals} Short LLM-generated clinical note summaries retained most signals necessary for downstream heart failure prediction. \Sidistill (almost 20x compression) maintained AUROC performance within 1.2–4.0\% of the full-note baseline, indicating that LLM summaries may replace full notes for some tasks while holding most functional performance, pointing towards, for example, real-time clinical risk modeling based on summaries, improving speed and scalability.

\paragraph{Different summarization strategies introduce different tradeoffs} While all summary types retained most of the diagnostic performance, highly compressed summaries (Distill) still resulted in slightly lower prediction performance metrics when compared to the least compressed (respectively, Structured and One-step), indicating the progressive loss of diagnostic signals compression brings. Making these tradeoffs explicit with \texttt{DistillNote} gives important insight for deployment: shorter summaries may still be very rich in diagnostic signals; however, small decreases in performance could pose clinical risks for certain groups of patients depending on their final use purpose.

\paragraph{Model choice affects diagnosis performance} There were slight differences in how general-purpose, biomedical, and reasoning LLMs predicted heart failure from summaries. Phi-4 14B (general) performed best with One-step and Structured summaries, while DeepSeek-R1 70B (reasoning) was the best for Distilled. OpenBioLLM, the biomedical model, has not outperformed in any approach. This suggests that smaller, general models may sometimes outperform in predicting complex clinical tasks; hence, model choice should be carefully evaluated before deployment.

\paragraph{Functional evaluation adds evaluation value} LLM-as-judge and clinician scores aligned with functional performance, but did not fully predict it. Distilled summaries were scored higher by them than others on factual accuracy, for example, but underperformed on \texttt{DistillNote}'s key prediction metrics. This discrepancy highlights the importance of introducing a third, task-based evaluation lens to help understand if summaries generated by a given LLM system may support important downstream clinical tasks, going beyond subjective assessment. 

\section{Conclusion}

We introduced \texttt{DistillNote}, an objective and domain-agnostic framework for assessing the quality of LLM-generated summaries by linking them to clinical prediction outcomes. We validated it by generating and evaluating over 192,000 LLM summaries for heart failure prediction. Models achieved significant text reduction while maintaining most prediction performance, highlighting tradeoffs that need to be carefully considered before deployment. By detailing the balance between compression and functional performance, \texttt{DistillNote} helps a more informed integration of LLM summarizers into clinical workflows.

Future work should explore \texttt{DistillNote} in other clinical tasks and notes, and examine its application in real-time clinical risk modeling.

%By linking LLM summary evaluation to clinical prediction outcomes, \texttt{DistillNote} provides a practical, objective, and medical domain-agnostic method to assess real-world usefulness. The framework, by making explicit compression-to-performance tradeoffs, can guide safer integration of LLM summarization systems into clinical workflows.

\section{Competing Interests}
The authors declare that they have no conflict of interest.

\section{Acknowledgments and funding}
HOB, and IC are funded by the project CaRe-NLP with file number NGF.1607.22.014 of the research programme AiNed Fellowship Grants which is (partly) financed by the Dutch Research Council (NWO).

\bibliographystyle{elsarticle-num} 
\bibliography{refs}

@Article{Johnson2023,
author={Johnson, Alistair E. W.
and Bulgarelli, Lucas
and Shen, Lu
and Gayles, Alvin
and Shammout, Ayad
and Horng, Steven
and Pollard, Tom J.
and Hao, Sicheng
and Moody, Benjamin
and Gow, Brian
and Lehman, Li-wei H.
and Celi, Leo A.
and Mark, Roger G.},
title={MIMIC-IV, a freely accessible electronic health record dataset},
journal={Scientific Data},
year={2023},
month={Jan},
day={03},
volume={10},
number={1},
pages={1},
issn={2052-4463},
doi={10.1038/s41597-022-01899-x},
url={https://doi.org/10.1038/s41597-022-01899-x}
}

@article{VanVeen:23,
	author = {D. Van Veen and C. Van Uden and L. Blankemeier and J.B. Delbrouck and A. Aali and C. Bluethgen and A. Pareek and M. Polacin and E.P. Reis and A. Seehofnerová and N. Rohatgi and P. Hosamani and W. Collins and N. Ahuja and C.P. Langlotz and J. Hom and S. Gatidis and J. Pauly and A.S. Chaudhari},
	year = "2023",
	title = {Clinical Text Summarization: Adapting Large Language Models Can Outperform Human Experts},
	journal = {Research Square [Preprint]},
	date = {2023-10-30},
	doi = {10.21203/rs.3.rs-3483777/v1},
	note = {Update in: Nat Med. 2024 Apr;30(4):1134-1142. doi: 10.1038/s41591-024-02855-5. PMID: 37961377; PMCID: PMC10635391}
}

@inproceedings{liang_novel_2019,
    title = "A Novel System for Extractive Clinical Note Summarization using {EHR} Data",
    author = "Liang, Jennifer  and
      Tsou, Ching-Huei  and
      Poddar, Ananya",
    editor = "Rumshisky, Anna  and
      Roberts, Kirk  and
      Bethard, Steven  and
      Naumann, Tristan",
    booktitle = "Proceedings of the 2nd Clinical Natural Language Processing Workshop",
    month = jun,
    year = "2019",
    address = "Minneapolis, Minnesota, USA",
    publisher = "Association for Computational Linguistics",
    url = "https://aclanthology.org/W19-1906/",
    doi = "10.18653/v1/W19-1906",
    pages = "46--54"
}

@misc{chuang_spec_2023,
	title = {{SPeC}: A Soft Prompt-Based Calibration on Performance Variability of Large Language Model in Clinical Notes Summarization},
	url = {http://arxiv.org/abs/2303.13035},
	doi = {10.48550/arXiv.2303.13035},
	shorttitle = {{SPeC}},
	number = {{arXiv}:2303.13035},
	publisher = {{arXiv}},
    year = {2023},
	author = {Chuang, Yu-Neng and Tang, Ruixiang and Jiang, Xiaoqian and Hu, Xia},
	urldate = {2025-01-30},
	date = {2023-08-04},
	eprinttype = {arxiv},
	eprint = {2303.13035 [cs]},
}

@inproceedings{rohr_revisiting_2024,
	location = {Mexico City, Mexico},
	title = {Revisiting Clinical Outcome Prediction for {MIMIC}-{IV}},
    year = {2024},
	url = {https://aclanthology.org/2024.clinicalnlp-1.18/},
	doi = {10.18653/v1/2024.clinicalnlp-1.18},
	pages = {208--217},
	booktitle = {Proceedings of the 6th Clinical Natural Language Processing Workshop},
	publisher = {Association for Computational Linguistics},
	author = {Röhr, Tom and Figueroa, Alexei and Papaioannou, Jens-Michalis and Fallon, Conor and Bressem, Keno and Nejdl, Wolfgang and Löser, Alexander},
	editor = {Naumann, Tristan and Ben Abacha, Asma and Bethard, Steven and Roberts, Kirk and Bitterman, Danielle},
	urldate = {2025-01-23},
	date = {2024-06},
}

@misc{abdin_phi-4_2024,
	title = {Phi-4 Technical Report},
	url = {http://arxiv.org/abs/2412.08905},
	doi = {10.48550/arXiv.2412.08905},
	number = {{arXiv}:2412.08905},
	publisher = {{arXiv}},
    year = {2024},
	author = {Abdin, Marah and Aneja, Jyoti and Behl, Harkirat and Bubeck, Sébastien and Eldan, Ronen and Gunasekar, Suriya and Harrison, Michael and Hewett, Russell J. and Javaheripi, Mojan and Kauffmann, Piero and Lee, James R. and Lee, Yin Tat and Li, Yuanzhi and Liu, Weishung and Mendes, Caio C. T. and Nguyen, Anh and Price, Eric and Rosa, Gustavo de and Saarikivi, Olli and Salim, Adil and Shah, Shital and Wang, Xin and Ward, Rachel and Wu, Yue and Yu, Dingli and Zhang, Cyril and Zhang, Yi},
	urldate = {2025-02-04},
	date = {2024-12-12},
	eprinttype = {arxiv},
	eprint = {2412.08905 [cs]},
}

@misc{bavaresco_llms_2024,
	title = {{LLMs} instead of Human Judges? A Large Scale Empirical Study across 20 {NLP} Evaluation Tasks},
	url = {http://arxiv.org/abs/2406.18403},
	doi = {10.48550/arXiv.2406.18403},
	shorttitle = {{LLMs} instead of Human Judges?},
    year = {2024},
	number = {{arXiv}:2406.18403},
	publisher = {{arXiv}},
	author = {Bavaresco, Anna and Bernardi, Raffaella and Bertolazzi, Leonardo and Elliott, Desmond and Fernández, Raquel and Gatt, Albert and Ghaleb, Esam and Giulianelli, Mario and Hanna, Michael and Koller, Alexander and Martins, André F. T. and Mondorf, Philipp and Neplenbroek, Vera and Pezzelle, Sandro and Plank, Barbara and Schlangen, David and Suglia, Alessandro and Surikuchi, Aditya K. and Takmaz, Ece and Testoni, Alberto},
	urldate = {2025-02-04},
	date = {2024-12-19},
	eprinttype = {arxiv},
	eprint = {2406.18403 [cs]},
}

@misc{krolik_towards_2024,
	title = {Towards Leveraging Large Language Models for Automated Medical Q\&A Evaluation},
    year = {2024},
	url = {http://arxiv.org/abs/2409.01941},
	doi = {10.48550/arXiv.2409.01941},
	number = {{arXiv}:2409.01941},
	publisher = {{arXiv}},
	author = {Krolik, Jack and Mahal, Herprit and Ahmad, Feroz and Trivedi, Gaurav and Saket, Bahador},
	urldate = {2025-02-04},
	date = {2024-09-03},
	eprinttype = {arxiv},
	eprint = {2409.01941 [cs]},
}

@misc{lee_distillation_2025,
	title = {Distillation Quantification for Large Language Models},
    year = {2025},
	url = {http://arxiv.org/abs/2501.12619},
	doi = {10.48550/arXiv.2501.12619},
	number = {{arXiv}:2501.12619},
	publisher = {{arXiv}},
	author = {Lee, Sunbowen and Zhou, Junting and Ao, Chang and Li, Kaige and Du, Xinrun and He, Sirui and Liu, Jiaheng and Yang, Min and Wen, Zhoufutu and Ni, Shiwen},
	urldate = {2025-02-05},
	date = {2025-01-30},
	eprinttype = {arxiv},
	eprint = {2501.12619 [cs]},
}

@misc{subramanian_small_2025,
	title = {Small Language Models ({SLMs}) Can Still Pack a Punch: A survey},
    year = {2025},
	url = {http://arxiv.org/abs/2501.05465},
	doi = {10.48550/arXiv.2501.05465},
	shorttitle = {Small Language Models ({SLMs}) Can Still Pack a Punch},
	number = {{arXiv}:2501.05465},
	publisher = {{arXiv}},
	author = {Subramanian, Shreyas and Elango, Vikram and Gungor, Mecit},
	urldate = {2025-02-05},
	date = {2025-01-03},
	eprinttype = {arxiv},
	eprint = {2501.05465 [cs]},
}

@misc{johnson_mimic-iv-note_nodate,
	title = {{MIMIC}-{IV}-Note: Deidentified free-text clinical notes},
	url = {https://physionet.org/content/mimic-iv-note/2.2/},
	doi = {10.13026/1N74-NE17},
	shorttitle = {{MIMIC}-{IV}-Note},
    year=2023,
	version = {2.2},
	publisher = {{PhysioNet}},
	author = {Johnson, Alistair and Pollard, Tom and Horng, Steven and Celi, Leo Anthony and Mark, Roger},
	urldate = {2025-02-05},
}

@article{fraile_navarro_expert_2025,
	title = {Expert evaluation of large language models for clinical dialogue summarization},
    year = {2025},
	volume = {15},
	rights = {2025 The Author(s)},
	issn = {2045-2322},
    journal = {Scientific Reports},
	url = {https://www.nature.com/articles/s41598-024-84850-x},
	doi = {10.1038/s41598-024-84850-x},
	pages = {1195},
	number = {1},
	journaltitle = {Scientific Reports},
	shortjournal = {Sci Rep},
	author = {Fraile Navarro, David and Coiera, Enrico and Hambly, Thomas W. and Triplett, Zoe and Asif, Nahyan and Susanto, Anindya and Chowdhury, Anamika and Azcoaga Lorenzo, Amaya and Dras, Mark and Berkovsky, Shlomo},
	urldate = {2025-02-06},
	date = {2025-01-07},
	langid = {english},
	note = {Publisher: Nature Publishing Group},
}

@article{chen_exploring_2024,
    year = {2024},
	title = {Exploring the opportunities of large language models for summarizing palliative care consultations: A pilot comparative study},
	volume = {10},
    journal = {Digital Health},
	issn = {2055-2076},
	url = {https://doi.org/10.1177/20552076241293932},
	doi = {10.1177/20552076241293932},
	shorttitle = {Exploring the opportunities of large language models for summarizing palliative care consultations},
	pages = {20552076241293932},
	journaltitle = {{DIGITAL} {HEALTH}},
	author = {Chen, Xiao and Zhou, Wei and Hoda, Rashina and Li, Andy and Bain, Chris and Poon, Peter},
	urldate = {2025-02-06},
	date = {2024-09-01},
	langid = {english},
	note = {Publisher: {SAGE} Publications Ltd},
}

@article{arokodare_clinical_2025,
  author = {Arokodare, O. and Wimmer, H. and Du, J.},
  title = {Clinical Text Summarization Using NLP Pretrained Language Models: A Case Study of MIMIC-IV-Notes},
  journal = {Journal of Information Systems Applied Research and Analytics},
  year = {2025},
  volume = {18},
  number = {1},
  pages = {17--31},
  doi = {10.62273/NAKA3054},
  url = {https://doi.org/10.62273/NAKA3054}
}

@article{li2024improving,
  title={Improving Clinical Note Generation from Complex Doctor-Patient Conversation},
  author={Yizhan Li and Sifan Wu and Christopher Smith and Thomas Lo and Bang Liu},
  journal={arXiv preprint arXiv:2408.14568},
  year={2024},
  url={https://doi.org/10.48550/arXiv.2408.14568},
  archivePrefix={arXiv},
  eprint={2408.14568},
  primaryClass={cs.CL}
}

@article{choudhuri2025summarizing,
  title={Summarizing Clinical Notes using LLMs for ICU Bounceback and Length-of-Stay Prediction},
  author={Akash Choudhuri and Philip Polgreen and Alberto Segre and Bijaya Adhikari},
  journal={medRxiv preprint},
  year={2025},
  doi={10.1101/2025.01.19.25320797},
  note={This article is a preprint and has not been peer-reviewed}
}

@article{Verma2025VerifiableSO,
  title={Verifiable Summarization of Electronic Health Records Using Large Language Models to Support Chart Review},
  author={Ritchie Verma and Emily Alsentzer and Zachary Strasser and Leslie Chang and Kirollos Roman and Esteban Gershanik and Camellia Hernandez and Miguel Linares and Jorge Rodriguez and Durga Thakral and Ozan Unlu and Jacqueline You and Li Zhou and David Bates},
  journal={medRxiv},
  year={2025},
  url={https://api.semanticscholar.org/CorpusID:279089299}
}

@article{jung2025enhancing,
  title={Enhancing Clinical Efficiency through LLM: Discharge Note Generation for Cardiac Patients},
  author={HyoJe Jung and Yunha Kim and Heejung Choi and Hyeram Seo and Minkyoung Kim and JiYe Han and Gaeun Kee and Seohyun Park and Soyoung Ko and Byeolhee Kim and Suyeon Kim and Tae Joon Jun and Young-Hak Kim},
  journal={Healthcare AI Research},
  year={2025},
  note={This study evaluates the use of Mistral-7B for automating discharge note generation for cardiac patients},
  url={https://doi.org/10.1234/healthcare-ai.2025.example-url}
}

@misc{hegselmann2024annotations,
  title={Medical Expert Annotations of Unsupported Facts in Doctor-Written and LLM-Generated Patient Summaries (version 1.0.0)},
  author={Hegselmann, S. and Shen, S. and Gierse, F. and Agrawal, M. and Sontag, D. and Jiang, X.},
  year={2024},
  publisher={PhysioNet},
  url={https://doi.org/10.13026/a66y-aa53}
}

@article{hegselmann2024datacentric,
  title={A Data-Centric Approach To Generate Faithful and High Quality Patient Summaries with Large Language Models},
  author={Hegselmann, S. and Shen, S. Z. and Gierse, F. and Agrawal, M. and Sontag, D. and Jiang, X.},
  journal={arXiv preprint arXiv:2402.15422},
  year={2024}
}

@misc{schoonbeek_completeness_2024,
	title = {Completeness, Correctness and Conciseness of Physician-Written Versus Large Language Model Generated Patient Summaries Integrated in Electronic Health Records},
	url = {https://www.ssrn.com/abstract=4835935},
	doi = {10.2139/ssrn.4835935},
	publisher = {{SSRN}},
	author = {Schoonbeek, Rosanne and Workum, Jessica and Schuit, Stephanie  C.E. and Doornberg, Job and Van Der Laan, Tom  P. and Bootsma-Robroeks, Charlotte  M.H.H.T.},
	urldate = {2025-02-14},
	year = {2024},
	langid = {english}
}

@misc{OpenBioLLM,
  author    = {Ankit Pal and Malaikannan Sankarasubbu},
  title     = {OpenBioLLMs: Advancing Open-Source Large Language Models for Healthcare and Life Sciences},
  year      = {2024},
  publisher = {Hugging Face},
  journal   = {Hugging Face repository},
  howpublished = {\url{https://huggingface.co/aaditya/OpenBioLLM-Llama3-70B}}
}

@article{grattafiori_llama3_2024,
  author    = {Aaron Grattafiori and Abhimanyu Dubey and Abhinav Jauhri and et al.},
  title     = {The Llama 3 Herd of Models},
  journal   = {arXiv preprint},
  year      = {2024},
  archivePrefix = {arXiv},
  eprint    = {2407.21783},
  primaryClass = {cs.AI},
  doi       = {10.48550/arXiv.2407.21783},
  url       = {https://doi.org/10.48550/arXiv.2407.21783}
}

@article{deepseek_r1_2025,
  author    = {DeepSeek-AI and Daya Guo and Dejian Yang and Haowei Zhang and et al.},
  title     = {DeepSeek-R1: Incentivizing Reasoning Capability in LLMs via Reinforcement Learning},
  journal   = {arXiv preprint},
  year      = {2025},
  archivePrefix = {arXiv},
  eprint    = {2501.12948},
  primaryClass = {cs.CL},
  doi       = {10.48550/arXiv.2501.12948},
  url       = {https://doi.org/10.48550/arXiv.2501.12948}
}

@misc{zheng_llamafactory_2024,
  author    = {Yaowei Zheng and Richong Zhang and Junhao Zhang and Yanhan Ye and Zheyan Luo and Zhangchi Feng and Yongqiang Ma},
  title     = {LlamaFactory: Unified Efficient Fine-Tuning of 100+ Language Models},
  booktitle = {Proceedings of the 62nd Annual Meeting of the Association for Computational Linguistics (Volume 3: System Demonstrations)},
  year      = {2024},
  publisher = {Association for Computational Linguistics},
  address   = {Bangkok, Thailand},
  url       = {https://arxiv.org/abs/2403.13372}
}

@inproceedings{rasley_deepspeed_2020,
  author    = {Jeff Rasley and Samyam Rajbhandari and Olatunji Ruwase and Yuxiong He},
  title     = {DeepSpeed: System Optimizations Enable Training Deep Learning Models with Over 100 Billion Parameters},
  booktitle = {Proceedings of the 26th ACM SIGKDD International Conference on Knowledge Discovery \& Data Mining},
  pages     = {3505--3506},
  year      = {2020},
  publisher = {ACM},
  doi       = {10.1145/3394486.3406703},
  url       = {https://doi.org/10.1145/3394486.3406703}
}

@article{chen2024predicting,
  title={Predicting in-hospital mortality in patients with heart failure combined with atrial fibrillation using stacking ensemble model: an analysis of the medical information mart for intensive care IV (MIMIC-IV)},
  author={Chen, Pei and Sun, Jiming and Chu, Yan and Jiang, Yue and Yuan, Luyu and Liang, Jian and Jiang, Chun and Jiang, Xiangyu and Shen, Hao and Xu, Fulin and Huang, Zhenyu},
  journal={BMC Medical Informatics and Decision Making},
  volume={24},
  number={1},
  pages={402},
  year={2024},
  publisher={BioMed Central},
  doi={10.1186/s12911-024-02829-0}
}

@article{hu2021lora,
  title={LoRA: Low-Rank Adaptation of Large Language Models},
  author={Hu, Edward J. and Shen, Yelong and Wallis, Phillip and Allen-Zhu, Zeyuan and Li, Yuanzhi and Wang, Shean and Wang, Lu and Chen, Weizhu},
  journal={arXiv preprint arXiv:2106.09685},
  year={2021},
  note={Version 2, revised October 16, 2021},
  url={https://doi.org/10.48550/arXiv.2106.09685}
}

@article{liu2023geval,
  title={G-Eval: NLG Evaluation using GPT-4 with Better Human Alignment},
  author={Yang Liu and Dan Iter and Yichong Xu and Shuohang Wang and Ruochen Xu and Chenguang Zhu},
  journal={arXiv preprint arXiv:2303.16634},
  year={2023},
  url={https://arxiv.org/abs/2303.16634}
}

@article{li_predicting_2022,
	title = {Predicting Mortality in Intensive Care Unit Patients With Heart Failure Using an Interpretable Machine Learning Model: Retrospective Cohort Study},
	volume = {24},
    journal = {Journal of Medical Internet Research},
	url = {https://www.jmir.org/2022/8/e38082},
	doi = {10.2196/38082},
	shorttitle = {Predicting Mortality in Intensive Care Unit Patients With Heart Failure Using an Interpretable Machine Learning Model},
    year = {2022},
	pages = {e38082},
	number = {8},
	journaltitle = {Journal of Medical Internet Research},
	author = {Li, Jili and Liu, Siru and Hu, Yundi and Zhu, Lingfeng and Mao, Yujia and Liu, Jialin},
	urldate = {2025-04-30},
	date = {2022-08-09},
	note = {Company: Journal of Medical Internet Research
Distributor: Journal of Medical Internet Research
Institution: Journal of Medical Internet Research
Label: Journal of Medical Internet Research
Publisher: {JMIR} Publications Inc., Toronto, Canada},
}

@article{li_machine_2025,
	title = {Machine Learning for In-hospital Mortality Prediction in Critically Ill Patients With Acute Heart Failure: A Retrospective Analysis Based on the {MIMIC}-{IV} Database},
	volume = {39},
	issn = {1053-0770},
    journal = {Journal of Cardiothoracic and Vascular Anesthesia},
	url = {https://www.sciencedirect.com/science/article/pii/S1053077024009819},
    year = {2025},
	doi = {10.1053/j.jvca.2024.12.016},
	shorttitle = {Machine Learning for In-hospital Mortality Prediction in Critically Ill Patients With Acute Heart Failure},
	pages = {666--674},
	number = {3},
	journaltitle = {Journal of Cardiothoracic and Vascular Anesthesia},
	shortjournal = {Journal of Cardiothoracic and Vascular Anesthesia},
	author = {Li, Jun and Sun, Yiwu and Ren, Jie and Wu, Yifan and He, Zhaoyi},
	urldate = {2025-04-30},
	date = {2025-03-01},
}

@article{o_risk_2023,
	title = {Risk Prediction for Heart Failure Patients Admitted to the Intensive Care Unit},
	volume = {11},
    year = {2023},
    journal = {JACC: Heart Failure},
	url = {https://www.jacc.org/doi/10.1016/j.jchf.2023.01.021},
	doi = {10.1016/j.jchf.2023.01.021},
	pages = {727--728},
	number = {6},
	journaltitle = {{JACC}: Heart Failure},
	author = {O, Connor Kyle D. and Yamamoto, Yu and Sen, Sounok and Samsky, Marc D. and Wilson, F. Perry and Desai, Nihar and Ahmad, Tariq and Fuery, Michael A.},
	urldate = {2025-04-30},
	date = {2023-06},
	note = {Publisher: American College of Cardiology Foundation},
}

@inproceedings{bavaresco-etal-2025-llms,
    title = "{LLM}s instead of Human Judges? A Large Scale Empirical Study across 20 {NLP} Evaluation Tasks",
    author = "Bavaresco, Anna  and
      Bernardi, Raffaella  and
      Bertolazzi, Leonardo  and
      Elliott, Desmond  and
      Fern{\'a}ndez, Raquel  and
      Gatt, Albert  and
      Ghaleb, Esam  and
      Giulianelli, Mario  and
      Hanna, Michael  and
      Koller, Alexander  and
      Martins, Andre  and
      Mondorf, Philipp  and
      Neplenbroek, Vera  and
      Pezzelle, Sandro  and
      Plank, Barbara  and
      Schlangen, David  and
      Suglia, Alessandro  and
      Surikuchi, Aditya K  and
      Takmaz, Ece  and
      Testoni, Alberto",
    editor = "Che, Wanxiang  and
      Nabende, Joyce  and
      Shutova, Ekaterina  and
      Pilehvar, Mohammad Taher",
    booktitle = "Proceedings of the 63rd Annual Meeting of the Association for Computational Linguistics (Volume 2: Short Papers)",
    month = jul,
    year = "2025",
    address = "Vienna, Austria",
    publisher = "Association for Computational Linguistics",
    url = "https://aclanthology.org/2025.acl-short.20/",
    doi = "10.18653/v1/2025.acl-short.20",
    pages = "238--255",
    ISBN = "979-8-89176-252-7"
}

@misc{gebauer_benchmarking_2025,
  author = {Gebauer, Sarah},
  title = {Benchmarking and Datasets for Ambient Clinical Documentation: A Scoping Review of Existing Frameworks and Metrics for AI-Assisted Medical Note Generation},
  year = {2025},
  doi = {10.1101/2025.01.29.25320859},
  eprint = {2025.01.29.25320859},
  archivePrefix = {medRxiv},
  url = {https://doi.org/10.1101/2025.01.29.25320859}
}

@inproceedings{NEURIPS2024_4df3510a,
 author = {McDermott, 
           
           Matthew B. and Zhang, Haoran and Hansen, Lasse Hyldig and Angelotti, Giovanni and Gallifant, Jack},
 booktitle = {Advances in Neural Information Processing Systems},
 editor = {A. Globerson and L. Mackey and D. Belgrave and A. Fan and U. Paquet and J. Tomczak and C. Zhang},
 pages = {44102--44163},
 publisher = {Curran Associates, Inc.},
 title = {A Closer Look at AUROC and AUPRC under Class Imbalance},
 url = {https://proceedings.neurips.cc/paper_files/paper/2024/file/4df3510ad02a86d69dc32388d91606f8-Paper-Conference.pdf},
 volume = {37},
 year = {2024}
}

@inproceedings{liang-etal-2019-novel-system,
    title = "A Novel System for Extractive Clinical Note Summarization using {EHR} Data",
    author = "Liang, Jennifer  and
      Tsou, Ching-Huei  and
      Poddar, Ananya",
    editor = "Rumshisky, Anna  and
      Roberts, Kirk  and
      Bethard, Steven  and
      Naumann, Tristan",
    booktitle = "Proceedings of the 2nd Clinical Natural Language Processing Workshop",
    month = jun,
    year = "2019",
    address = "Minneapolis, Minnesota, USA",
    publisher = "Association for Computational Linguistics",
    url = "https://aclanthology.org/W19-1906/",
    doi = "10.18653/v1/W19-1906",
    pages = "46--54"
}

@article{goh2024large,
  title     = {Large Language Model Influence on Diagnostic Reasoning: A Randomized Clinical Trial},
  author    = {Goh, E and Gallo, R and Hom, J and others},
  journal   = {JAMA Network Open},
  year      = {2024},
  volume    = {7},
  number    = {10},
  pages     = {e2440969},
  doi       = {10.1001/jamanetworkopen.2024.40969}
}

@inproceedings{vanSchaikPugh24,
  author    = {Tempest A. van Schaik and Brittany Pugh},
  title     = {A Field Guide to Automatic Evaluation of LLM-Generated Summaries},
  booktitle = {Proceedings of the 47th International ACM SIGIR Conference on Research and Development in Information Retrieval (SIGIR ’24)},
  pages     = {2832–2836},
  year      = {2024},
  doi       = {10.1145/3626772.3661346},
  url       = {https://doi.org/10.1145/3626772.3661346}
}

@misc{Jabbour2025EvaluationFF,
  title={Evaluation Framework for AI Systems in"the Wild"},
  author={Sarah Jabbour and Trenton Chang and Anindya Das Antar and Joseph Peper and Insu Jang and Jiachen Liu and Jae-Won Chung and Shiqi He and Michael P. Wellman and Bryan Goodman and Elizabeth Bondi-Kelly and Kevin Samy and Rada Mihalcea and Mosharaf Chowhury and David Jurgens and Lu Wang},
  year={2025},
  url={https://api.semanticscholar.org/CorpusID:278000613}
}

@article{croxford_current_2025,
  author = {Croxford, E. and Gao, Y. and Pellegrino, N. and Wong, K. and Wills, G. and First, E. and Liao, F. and Goswami, C. and Patterson, B. and Afshar, M.},
  title = {Current and Future State of Evaluation of Large Language Models for Medical Summarization Tasks},
  journal = {NPJ Health Systems},
  year = {2025},
  volume = {2},
  pages = {6},
  doi = {10.1038/s44401-024-00011-2},
  pmid = {40124388},
  pmcid = {PMC11928168},
  url = {https://doi.org/10.1038/s44401-024-00011-2},
  note = {Epub 2025 Feb 3}
}

@misc{schroeder_can_2024,
  author = {Schroeder, Kayla and Wood-Doughty, Zach},
  title = {Can You Trust LLM Judgments? Reliability of LLM-as-a-Judge},
  year = {2024},
  eprint = {2412.12509},
  archivePrefix = {arXiv},
  primaryClass = {cs.CL},
  url = {https://doi.org/10.48550/arXiv.2412.12509}
}

@article{tajirian2020influence,
 title={The Influence of Electronic Health Record Use on Physician Burnout: Cross-Sectional Survey},
 author={Tajirian, Talar and Stergiopoulos, Vicky and Strudwick, Gillian and Sequeira, Lydia and Sanches, Mikaela and Kemp, Jacqueline and Ramamoorthi, Karthik and Zhang, Tao and Jankowicz, Damian},
 journal={Journal of Medical Internet Research},
 volume={22},
 number={7},
 pages={e19274},
 year={2020},
 publisher={JMIR Publications Inc.},
 doi={10.2196/19274},
 pmid={32673234},
 pmcid={PMC7392132}
}

@article{nijor2022patient,
 title={Patient Safety Issues From Information Overload in Electronic Medical Records},
 author={Nijor, Shyam and Rallis, George and Lad, Nishant and Gokcen, Emin},
 journal={Journal of Patient Safety},
 volume={18},
 number={6},
 pages={e999--e1003},
 year={2022},
 publisher={Lippincott Williams \& Wilkins},
 doi={10.1097/PTS.0000000000001002},
 pmid={35985047},
 pmcid={PMC9422765}
}

@article{steinkamp2022prevalence,
 title={Prevalence and Sources of Duplicate Information in the Electronic Medical Record},
 author={Steinkamp, Jackson and Kantrowitz, Jonathan J and Airan-Javia, Suhani},
 journal={JAMA Network Open},
 volume={5},
 number={9},
 pages={e2233348},
 year={2022},
 publisher={American Medical Association},
 doi={10.1001/jamanetworkopen.2022.33348},
 pmid={36156143},
 pmcid={PMC9513649}
}

@ARTICLE{2020SciPy-NMeth,
  author  = {Virtanen, Pauli and Gommers, Ralf and Oliphant, Travis E. and
            Haberland, Matt and Reddy, Tyler and Cournapeau, David and
            Burovski, Evgeni and Peterson, Pearu and Weckesser, Warren and
            Bright, Jonathan and {van der Walt}, St{\'e}fan J. and
            Brett, Matthew and Wilson, Joshua and Millman, K. Jarrod and
            Mayorov, Nikolay and Nelson, Andrew R. J. and Jones, Eric and
            Kern, Robert and Larson, Eric and Carey, C J and
            Polat, {\.I}lhan and Feng, Yu and Moore, Eric W. and
            {VanderPlas}, Jake and Laxalde, Denis and Perktold, Josef and
            Cimrman, Robert and Henriksen, Ian and Quintero, E. A. and
            Harris, Charles R. and Archibald, Anne M. and
            Ribeiro, Ant{\^o}nio H. and Pedregosa, Fabian and
            {van Mulbregt}, Paul and {SciPy 1.0 Contributors}},
  title   = {{{SciPy} 1.0: Fundamental Algorithms for Scientific
            Computing in Python}},
  journal = {Nature Methods},
  year    = {2020},
  volume  = {17},
  pages   = {261--272},
  adsurl  = {https://rdcu.be/b08Wh},
  doi     = {10.1038/s41592-019-0686-2},
}

@article{Vallat2018,
    title = {Pingouin: statistics in Python},
    volume = {3},
    DOI = {10.21105/joss.01026},
    number = {31},
    journal = {Journal of Open Source Software},
    publisher = {The Open Journal},
    author = {Vallat,  Raphael},
    year = {2018},
    month = nov,
    pages = {1026}
}

@article{subramanian2025clinical,
  author    = {Subramanian, C. Raghu and Rosner, Benjamin I.},
  title     = {Advancing Toward Clinical Deployment of AI-Generated Discharge Summaries Beyond the Bench},
  journal   = {JAMA Network Open},
  year      = {2025},
  volume    = {8},
  number    = {8},
  pages     = {e2526350},
  doi       = {10.1001/jamanetworkopen.2025.26350}
}

@misc{epic2025ai,
  author       = {Epic},
  title        = {AI for Clinicians},
  year         = {2025},
  howpublished = {\url{https://www.epic.com/software/ai-clinicians/}},
  note         = {Accessed 2025-09-16}
}

@misc{epic2023generative,
  author       = {Epic},
  title        = {Cool Stuff Now: Epic and Generative AI},
  year         = {2023},
  howpublished = {\url{https://www.epic.com/epic/post/cool-stuff-now-epic-and-generative-ai/}},
  note         = {Accessed 2025-09-16}
}

@article{silberlust2025ai,
  author    = {Silberlust, J. and Solanki, P. and Stevens, E. R. and Genes, N. and Lim, E. and Sun, K. and Lewis, M. and Testa, P. and Szerencsy, A.},
  title     = {Artificial intelligence-generated encounter summaries: early insights from ambulatory clinicians at a large academic health system},
  journal   = {JAMIA Open},
  year      = {2025},
  volume    = {8},
  number    = {5},
  pages     = {ooaf096},
  doi       = {10.1093/jamiaopen/ooaf096},
  pmid      = {40904519},
  pmcid     = {PMC12403210},
  publisher = {Oxford University Press}
}
%\end{linenumbers}


\newpage
\section{Supplementary Materials}
\label{sec:appendix}

\subsection{Related work}

Other researchers have explored clinical summarization using LLMs. \cite{choudhuri2025summarizing} and \cite{jung2025enhancing, arokodare_clinical_2025} summarized discharge and progress notes with LLMs. \cite{fraile_navarro_expert_2025, li2024improving, chen_exploring_2024} focused on producing LLM summaries from medical conversations. Divide-and-conquer summarization was also explored in \cite{liang-etal-2019-novel-system}, however in an non-LLM, extractive setting. In our study, we generate summaries with LLMs from patient admission notes.

In terms of LLM summary quality, both model-based and human evaluations were studied. \cite{VanVeen:23, schoonbeek_completeness_2024} explored how LLM-generated summaries often outperformed those of human experts in completeness, conciseness, and correctness. \cite{chuang_spec_2023} proposed a model-agnostic calibration method; while \cite{hegselmann2024annotations, hegselmann2024datacentric} introduced strategies to reduce hallucinations. \cite{goh2024large} investigated how LLMs assisted physicians in diagnostic reasoning through a human-in-the-loop approach. \cite{subramanian2025clinical} showed that LLM summaries maintained clinician accuracy in answering questions about heart failure patients in a small manual evaluation. \texttt{DistillNote} introduces a new perspective to summary quality evaluation, measuring diagnostic signal retention across summaries with different compression rates for a relevant and complex downstream clinical task.

\subsection{Admission note details}
\label{appendix:admnotestats}

Admission notes provide initial patient assessments in the hospital, including chief complaints, medical history, and initial treatment plans. In our study, we focus on heart failure patients admitted to the intensive care unit (ICU), aligning with recent studies \cite{li_predicting_2022, li_machine_2025, o_risk_2023}.

As per \cite{rohr_revisiting_2024}, we truncate discharge notes into admission notes, keeping only sections available at admission time. The reader can find the statistics of each section in \autoref{tab:wordcount_stats}. For heart failure diagnosis, we use the labels from \cite{chen2024predicting}.

The dataset we used to generate the note summaries is MIMIC-IV, which consists of de-identified EHR information, and is available under a credentialed data use agreement via the Physionet portal. Our summaries are released under the PhysioNet Credentialed Health Data license and distributed solely for research purposes.

\begin{table}[t!]
\centering
\small
\resizebox{0.7\textwidth}{!}{%
\begin{tabular}{lrr}
\toprule
\textbf{Section} & \textbf{Avg word/section} & \textbf{Std} \\
\midrule
Chief Complaint & 7.83 & 34.21 \\
Present Illness & 193.55 & 130.63 \\
Medical History & 53.38 & 82.76 \\
Medication Administered & 64.50 & 63.39 \\
Allergies & 6.32 & 5.85 \\
Physical Exam & 75.08 & 54.82 \\
Family History & 13.40 & 14.52 \\
Social History & 0.02 & 3.41 \\
\bottomrule
\end{tabular}}
\caption{Statistics by admission note sections.}
\label{tab:wordcount_stats}
\end{table}

\subsection{Summarization prompts}
\label{prompts}

Each note \( A_i \in \mathcal{M} \) is processed independently four times, with each concatenated to a different summarization prompt strategy.  In the four \Sistruct prompts, we use a one-shot methodology, providing one example of the desired output format to give the LLM clear guidance on the structure and content of the target information. This aims to reduce task ambiguity, preventing the lack of structure in a zero-shot approach. In the \Sidistill prompt, we instruct the LLM to synthesize the four previous \Sistruct summaries into one final summary. The \Sionestep prompt follows a similar structure, but instead instructs the LLM to summarize the full original admission note at once. Below the reader can find the prompts utilized for generation across approaches.

\subsubsection{`One-step' prompt}

\begin{tcolorbox}[colback=gray!5, colframe=black!40!white, listing only, listing options={basicstyle=\ttfamily\footnotesize, breaklines=true}]
Summarize the patient's admission note, focusing on clinically relevant aspects affecting diagnosis, treatment, and risk assessment. For the synthesis, consider only explicitly documented information to maintain accuracy. Patient admission note: {\textbf{note}}
\end{tcolorbox}

\subsubsection{`Structured' prompts}

\begin{tcolorbox}[title=Chief complaint, colback=gray!5, colframe=black!40!white, listing only, listing options={basicstyle=\ttfamily\footnotesize, breaklines=true}]
Summarize the patient's primary reason for admission in one concise sentence based strictly on the chief complaint and history of present illness. Outputting more than one sentence or adding remarks or notes is strictly forbidden. Extract only explicitly documented information to maintain accuracy. Example output: 'The patient, a female, presented with worsening shortness of breath and lower extremity edema over three days.' Patient admission note: {\textbf{note}}
\end{tcolorbox}

\begin{tcolorbox}[title=Past medical history, colback=gray!5, colframe=black!40!white, listing only, listing options={basicstyle=\ttfamily\footnotesize, breaklines=true}]
Summarize the patient's past medical history in one concise sentence, including chronic conditions, major illnesses, hospitalizations, and surgeries. Only mention ongoing medications if recently changed and allergies if clinically relevant. Outputting more than one sentence or adding remarks or notes is strictly forbidden. Extract only explicitly documented information to maintain accuracy. Example output: 'The patient, a male, has a history of hypertension, type 2 diabetes, and a prior myocardial infarction with stent placement.' Patient admission note: {\textbf{note}}
\end{tcolorbox}

\begin{tcolorbox}[title=Physical exam, colback=gray!5, colframe=black!40!white, listing only, listing options={basicstyle=\ttfamily\footnotesize, breaklines=true}]
Summarize key physical exam findings in one concise sentence, including only significant abnormalities and pertinent negatives. Only include vital signs if explicitly relevant. Outputting more than one sentence or adding remarks or notes is strictly forbidden. Extract only explicitly documented information to maintain accuracy. Example output: 'The patient, a female, is afebrile with a distended abdomen, shifting dullness, and trace lower extremity edema.' Patient admission note: {\textbf{note}}
\end{tcolorbox}

\begin{tcolorbox}[title=Social/family history, colback=gray!5, colframe=black!40!white, listing only, listing options={basicstyle=\ttfamily\footnotesize, breaklines=true}]
Summarize relevant family history, social determinants, and lifestyle factors in one concise sentence, focusing only on hereditary risks, substance use, living situation, occupational exposures, and support systems. Outputting more than one sentence or adding remarks or notes is strictly forbidden. Extract only explicitly documented information to maintain accuracy. Example output: 'The patient, a male, lives alone, has a history of heavy alcohol use, and has a father with a history of early-onset cardiovascular disease.' Patient admission note: {\textbf{note}}
\end{tcolorbox}

\subsubsection{`Distilled' prompt}

\begin{tcolorbox}[colback=gray!5, colframe=black!40!white, listing only, listing options={basicstyle=\ttfamily\footnotesize, breaklines=true}]
Summarize the summaries extracted from the patient's admission note into a single cohesive admission summary. Focus on clinically relevant aspects affecting diagnosis, treatment, and risk assessment. For the synthesis, consider only explicitly documented information to maintain accuracy. Patient summaries: {\textbf{structured summary}}
\end{tcolorbox}

\subsection{Sampling parameters}
\label{samplingparams}
For summary generation, all models used the same decoding configuration across prompt types: \texttt{temperature}: 0.0,  \texttt{top\_p}: 0.9, \texttt{repetition\_penalty}: 1.2. The \texttt{max\_tokens} parameter varied:
\begin{itemize}
    \item Phi-4, OpenBioLLM: 300 for \Sistruct, 700 for \Sionestep and \Sidistill
    \item Deepseek-R1: 2000  for all prompt types
\end{itemize}
Outputs from Deepseek include internal reasoning markers between \texttt{<think>} tags preceding the final summary, requiring the generation of more tokens. Hence, we parsed and isolated the main summary content after generation.

\subsection{Summary details}

We conducted preliminary fine-tuning evaluations to identify the most promising LLM for each summarization approach based on the validation F1-score. Based on that, we selected Deepseek for generating \Sidistill summaries, and Phi4 for \Sistruct and \Sionestep summaries. The statistics of the generated summaries, in comparison with the original admission notes, are present in \autoref{tab:input_length_by_strategy}.

\begin{table}[t!]
\centering
\begin{tabular}{lcccccc}
\toprule
\textbf{Strategy} & \textbf{Model} & \textbf{Avg word} & \textbf{Std} & \textbf{Min} & \textbf{Max} & \textbf{Compr. (\%)} \\
\midrule
One-go     & Phi-4     & 262.04  & 41.18   & 106   & 438   & 36.4\% \\
Structured & Phi-4     & 194.65  & 39.69   & 74    & 482   & 52.8\% \\
Distilled  & Deepseek  & 86.51   & 20.02   & 25    & 248   & 79.0\% \\
Baseline   & N/A       & 412.12  & 196.74  & 18    & 2425  & 0.0\%  \\
\bottomrule
\end{tabular}
\caption{Statistics across summarization strategies and baseline case (original admission notes). 'Model' refers to the LLM used to generate each summary type. Compression is calculated relative to the average word count of the baseline.}
\label{tab:input_length_by_strategy}
\end{table}

\subsection{Fine-tuning}
\label{finetuning}

We fine-tuned models using LoRA \cite{hu2021lora} with LoRA+ enhancement for parameter efficiency. Training utilized DeepSpeed ZeRO-3 \cite{rasley_deepspeed_2020} for memory optimization with bf16 precision in NVIDIA H100 GPUs (4 for Deepseek and OpenBioLLM, 2 for Phi-4). We trained LLMs for one epoch using a cosine learning rate schedule (initial lr=5e-5, warmup=10\%) and evaluated models every 80 steps, selecting the best checkpoints based on validation performance. The fine-tuning of 1 model took around 1 day. The training implementation used LLaMA-Factory \cite{zheng_llamafactory_2024} with the Alpaca prompt template. Dataset statistics are available in \autoref{tab:splits}. We performed F1 threshold tuning based on validation sets and report the tuned F1s of the test sets in \autoref{tab:preds}.

\begin{table}[t!]
\small
\centering
\caption{Dataset split and class distribution.}
\label{tab:class_distribution}
\begin{tabular}{lccc}
\toprule
\textbf{Split} & \textbf{Total Samples} & \textbf{\% pos} & \textbf{\% neg} \\
\midrule
Train & 46,702 & 21.83\% & 78.17\% \\
Validation & 8,217 & 22.08\% & 77.92\% \\
Test & 9,815 & 21.67\% & 78.33\% \\
\midrule
\textbf{Total} & 64,734 & \textbf{21.86\%} & \textbf{78.14\%} \\
\bottomrule
\end{tabular}
\label{tab:splits}
\end{table}

The prompt we utilized for heart failure prediction can be found below. We note that, when predicting for the baseline case, we change "\textit{admission note summary}" to "\textit{admission note}".

\begin{tcolorbox}[colback=gray!5, colframe=black!40!white, listing only, listing options={basicstyle=\ttfamily\footnotesize, breaklines=true}] You are an expert in clinical diagnosis. Determine whether the patient had heart failure during this visit based on their admission note summary. Output only a number between double brackets: [[0]] for No, [[1]] for Yes. Patient summary: {\textbf{summary}}
\end{tcolorbox}

\subsubsection{Summary results}

\autoref{tab:tradeoff} details the performance of the three \Sionestep, \Sistruct, and \Sidistill summarization strategies by comparing their average word count, compression percentage, and the performance with respect to the F1, AUROC, and AUPRC of the best baseline (original note). The efficiency ratio, calculated as \% compression divided by metric drop, indicates the trade-off between text distillation and performance. A higher ratio indicates a more efficient summary, i.e. a summary with a greater reduction in length and relatively less impact on performance.

\begin{table}[t!]
\centering
\begin{tabular}{lcccccc}
\toprule
\textbf{Strategy} & \textbf{Avg words} & \textbf{ Avg compression} & \textbf{↓AUROC} & \textbf{↓AUPRC} & \textbf{↓F1} \\
\midrule
Baseline & 412 & --- & --- & --- & --- \\
One-step & 262 & 36.4\% & 1.2\% & 3.4\% & 3.7\% \\
Structured & 195 & 52.8\% & 1.8\% & 5.2\% & 5.1\% \\
Distilled & 87 & 79.0\% & 4.0\% & 7.7\% & 7.2\% \\
\bottomrule
\end{tabular}
\caption{Trade-off between performance metrics and text compression for different summarization strategies. Losses are calculated as the average performance decrease per strategy relative to the baseline average (AUROC=0.938, AUPRC=0.849, F1=0.769).} %Efficiency ratio = compression\%/AUROC loss\% (One-step: 30.3×, Structured: 29.3×, Distilled: 19.8×).}
\label{tab:tradeoff}
\end{table}

\begin{table}[t!]
\centering
\begin{tabular}{lcccc}
\toprule
\textbf{Strategy} & \textbf{Avg compression} & \textbf{AUROC eff.} & \textbf{AUPRC eff.} & \textbf{F1 eff.} \\
\midrule
One-step & 36.4\% & 30.3× & 10.7× & 9.9× \\
Structured & 52.8\% & 29.3× & 10.2× & 10.3× \\
Distilled & 79.0\% & 19.8× & 10.3× & 11.0× \\
\bottomrule
\end{tabular}
\caption{Compression efficiency ratios across performance metrics (compression\%/loss\%). AUROC demonstrates high efficiency (19.8-30.3×), enabling substantial text reduction with minimal discrimination loss. AUPRC shows consistent efficiency (~10×) while F1 exhibits moderate efficiency (6.0-7.2×), indicating compression-to-performance trade-offs.}
\label{tab:efficiency}
\end{table}

%\begin{table}[t!]
%\tiny
%\centering
%\begin{tabular}{lccccc}
%\toprule
%\textbf{Approach} & \textbf{p}\textsubscript{\tiny AUROC} & \textbf{p}\textsubscript{\tiny AUPRC} & \textbf{p}\textsubscript{\tiny F1} & \textbf{Equivalent} \\
%\midrule
%Conquer & 0.026* & 0.133 & 0.650 & 33.3\% \\
%Divide  & 0.000* & 0.001* & 0.297 & 66.7\% \\
%One-go  & 0.000* & 0.001* & 0.034* & 100\% \\
%\bottomrule
%\end{tabular}
%\caption{Statistical equivalence analysis using Two One-Sided Tests (TOST) with 10\% equivalence margin relative to baseline performance. *Significant equivalence (p < 0.05). Equivalent percentage represents the proportion of metrics achieving statistical equivalence. One-go approach demonstrates complete statistical equivalence across all performance metrics.}
%\label{tab:equivalence}
%\end{table}

\subsection{LLM-as-judge} 
\label{appendix:llm-judge}
In the LLM-as-judge step, we evaluate the quality of the generated summaries. We select Phi-4 as the judge model due to its improved alignment with human preferences \cite{abdin_phi-4_2024} and strong reasoning capacity despite its medium size (14B parameters). This cost-effective performance is important given our need to evaluate over 64,000 summaries across 3 metrics and 3 summary strategies.
For the methodology, like G-Eval \cite{liu2023geval}, we account for model uncertainty by adjusting the obtained scores based on token probability distributions. This helps address model uncertainty and smoothes out extreme scores when the model shows lower confidence.

We use continuous scoring to allow for a more nuanced evaluation. Instead of directly extracting e.g "4.2" from model output, we analyze the underlying probability distribution across top-k possible digit tokens. For example, for a model generating a score with the first digit $d_1$ and second digit $d_2$, we extract probabilities $p(d_1)$ and $p(d_2)$, along with the probabilities of the top-k (k = 5) alternative numbers the model considered. We then calculate a weighted average of all possible score combinations, where the weight is the product of the digit probabilities. The final score is calculated as:

\begin{equation}
\text{Score} = \frac{\sum_{d_1 \in D_1} \sum_{d_2 \in D_2} p(d_1) \cdot p(d_2) \cdot \text{val}(d_1.d_2)}{\sum_{d_1 \in D_1} \sum_{d_2 \in D_2} p(d_1) \cdot p(d_2)}
\end{equation}

where $D_1$ and $D_2$ are the sets of possible digits at each position and \text{val}()
converts the digit concatenation to its numerical value. 

\begin{figure}[t!]
    \centering
    \includegraphics[width=\columnwidth]{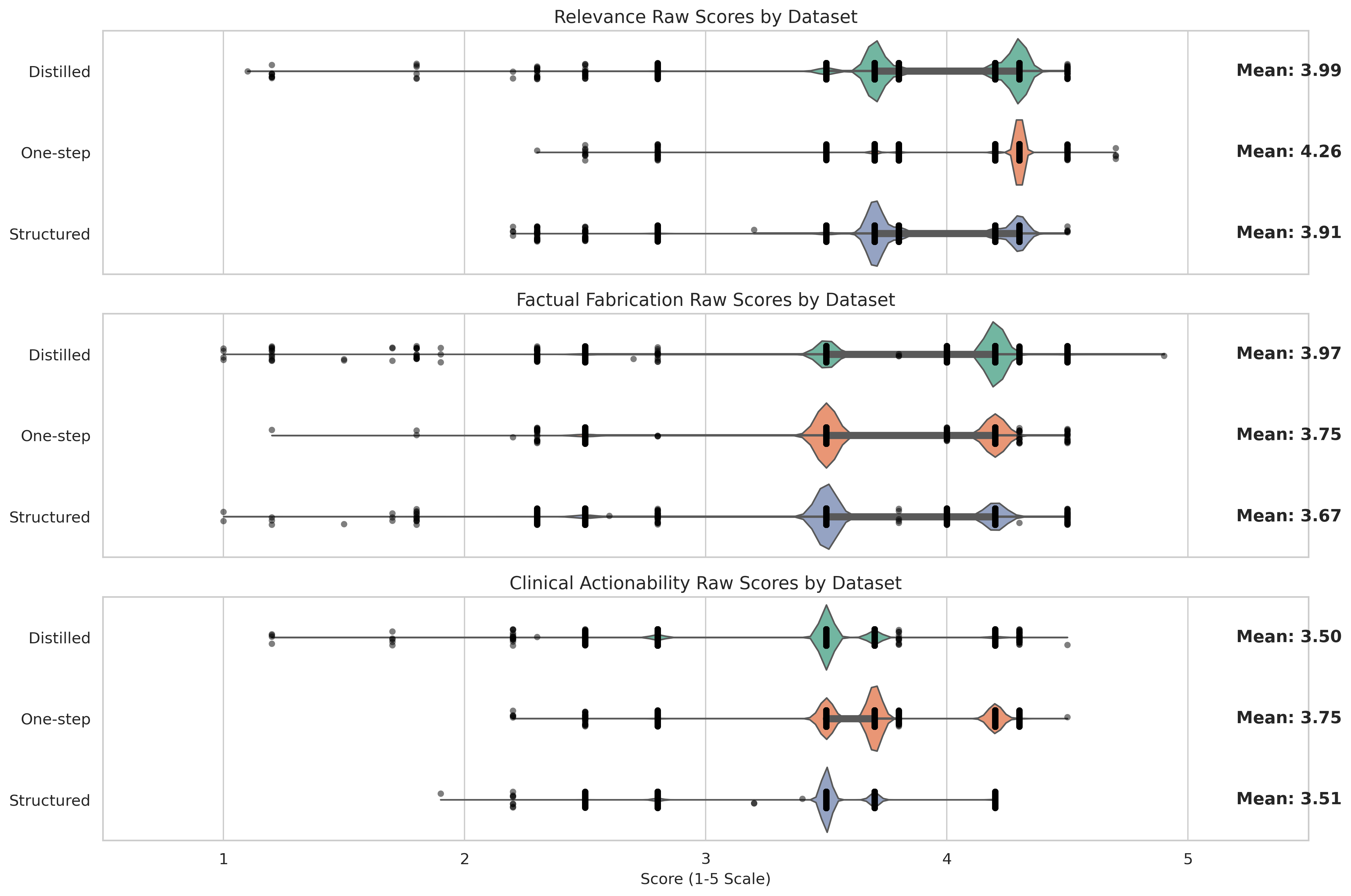}
    \caption{LLM-as-judge raw scores per summarization approach and metric.}
    \label{fig:raw-scores}
\end{figure}

\begin{figure}[t!]
    \centering
    \includegraphics[width=\columnwidth]{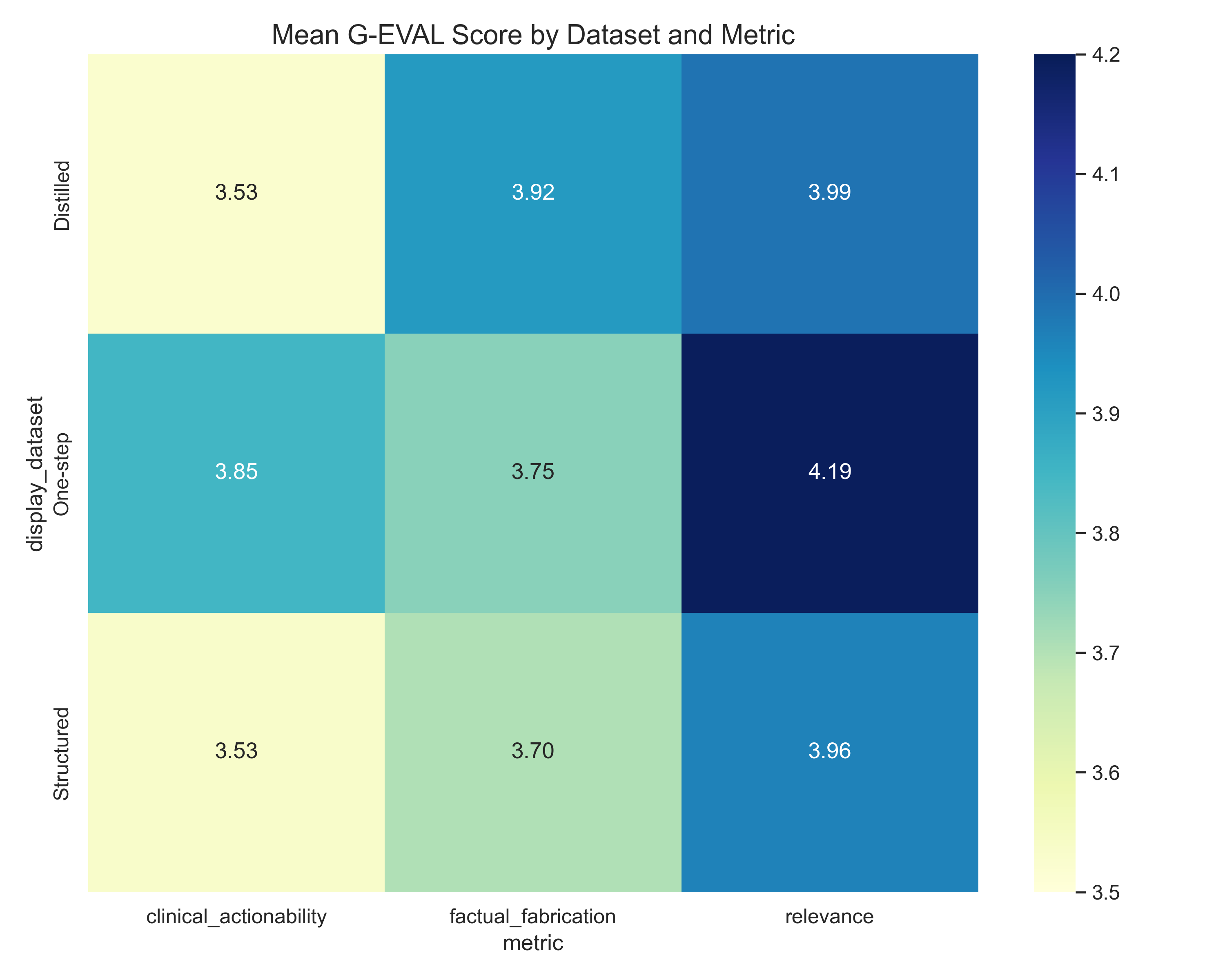}
    \caption{Heatmap of scores per summarization approach and metric.}
    \label{fig:heatmap-scores}
\end{figure}

\begin{figure}[t!]
    \centering
    \includegraphics[width=\columnwidth]{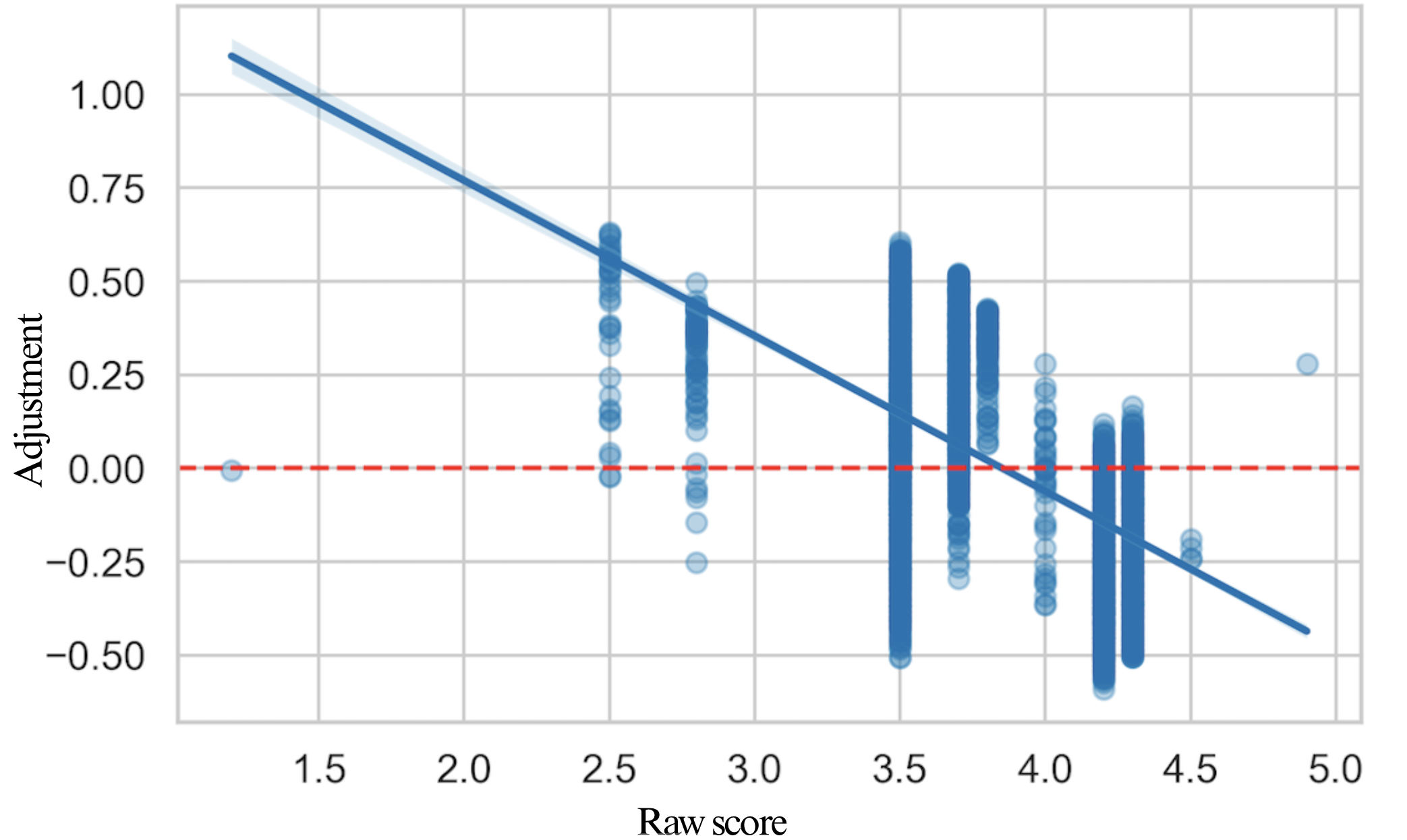}
    \caption{Score correction.}
    \label{fig:score_adjustments}
\end{figure}

The reader can find the distribution of original scores in \autoref{fig:raw-scores}, and the adjusted scores per summarization approach in \autoref{tab:geval-scores} and \autoref{fig:heatmap-scores}. \autoref{fig:score_adjustments} illustrates the adjustments made.

\begin{table}[t!]
\centering
\caption{LLM-as-judge scores by summary approach and metric (mean ± SD).}
\label{tab:geval-scores}
\begin{tabular}{lccc|c}
\toprule
\textbf{Approach} & \textbf{Fabrication} & \textbf{Relevance} & \textbf{Actionability} & \textbf{Overall} \\
\midrule
One-step & 3.75±0.28 & \textbf{4.19±0.15} & \textbf{3.85±0.22} & \textbf{3.93±0.29} \\
Structured & 3.70±0.31 & 3.96±0.18 & 3.53±0.24 & 3.73±0.30 \\
Distilled & \textbf{3.92±0.26} & 3.99±0.20 & 3.53±0.28 & 3.81±0.32 \\
\bottomrule
\end{tabular}
\end{table}

\subsection{LLM-as-judge score analysis}
\label{stats-llmjudge}
We applied four statistical tests with Scipy \cite{2020SciPy-NMeth} and Pingouin \cite{Vallat2018} to assess differences among summarization strategies. A one-way ANOVA (\autoref{tab:anova_dataset}) tested for significant differences between the \Sionestep, \Sistruct, and \Sidistill{} methods across the three metrics: relevance, factual fabrication, and clinical actionability. A two-way ANOVA (\autoref{tab:anova_interaction}) examined interaction effects between method and metric. Post hoc Tukey HSD tests (\autoref{tab:tukey}) identified specific pairwise differences, and Cohen’s d quantified effect sizes (\autoref{tab:effect_sizes}) to indicate the magnitude of the differences.

\begin{table}[t!]
\centering
\begin{tabular}{lccccc}
\toprule
\textbf{Source} & \textbf{SS} & \textbf{DF} & \textbf{MS} & \textbf{F} & \textbf{p-value} \\
\midrule
Dataset & 3820.81 & 2 & 1910.41 & 20529.63 & $<$0.001 \\
Residual & 54212.89 & 582583 & 0.093 & -- & -- \\
\bottomrule
\end{tabular}
\caption{One-way ANOVA testing for differences across summarization strategies. SS: Sum of Squares, DF: Degrees of Freedom, MS: Mean Square, F: F-statistic.}
\label{tab:anova_dataset}
\end{table}

\begin{table}[t!]
\centering
\begin{tabular}{lccccc}
\toprule
\textbf{Source} & \textbf{SS} & \textbf{DF} & \textbf{MS} & \textbf{F} & \textbf{p-value} \\
\midrule
Dataset & 3820.78 & 2 & 1910.39 & 33623.26 & $<$0.001 \\
Metric & 16843.69 & 2 & 8421.85 & 148226.31 & $<$0.001 \\
Interaction & 4268.63 & 4 & 1067.16 & 18782.21 & $<$0.001 \\
Residual & 33100.56 & 582577 & 0.057 & -- & -- \\
\bottomrule
\end{tabular}
\caption{Two-way ANOVA testing main and interaction effects of summarization strategy and metric on scores. SS: Sum of Squares, DF: Degrees of Freedom, MS: Mean Square, F: F-statistic.}
\label{tab:anova_interaction}
\end{table}

\begin{table}[t!]
\centering
\begin{tabular}{l l r r r r r r}
\toprule
\textbf{A} & \textbf{B} & \textbf{Diff} & \textbf{SE} & \textbf{T} & \textbf{p} & \textbf{Hedges' $g$} \\
\midrule
Distilled & One-step & -0.120 & 0.00098 & -122.69 & $<$0.001 & -0.39 \\
Distilled & Structured & 0.077 & 0.00098 & 78.31 & $<$0.001 & 0.25 \\
One-step & Structured & 0.197 & 0.00098 & 201.00 & $<$0.001 & 0.66 \\
\bottomrule
\end{tabular}
\caption{Tukey HSD post hoc test comparing summarization strategies. All differences are statistically significant.}
\label{tab:tukey}
\end{table}

\begin{table}[t!]
\small
\centering
\begin{tabular}{l r}
\toprule
\textbf{Comparison} & \textbf{Cohen's $d$} \\
\midrule
Distilled vs One-step & -0.39 \\
Distilled vs Structured & 0.25 \\
One-step vs Structured & 0.66 \\
\bottomrule
\end{tabular}
\caption{Cohen’s $d$ overall effect sizes between summarization strategies, indicating the magnitude of performance differences.}
\label{tab:effect_sizes}
\end{table}

\subsection{Clinical validation}

Doctors (n = 2) were recruited voluntarily. \autoref{tab:clinician_preferences} shows their preferences  between different summary approaches for 18 admission notes. Values represent the number of times each method was preferred. P-values were calculated using two-sided binomial tests with Bonferroni correction.

We calculated a Spearman correlation between clinician preferences and LLM-as-judge evaluations, which revealed a positive relationship (\autoref{tab:llm_clinician}).

\begin{table}[t!]
\centering
\begin{tabular}{llccccc}
\toprule
\textbf{Comparison} & \textbf{Metric} & \multicolumn{3}{c}{\textbf{Preference Count}} & \textbf{p-value} & \textbf{Significant} \\
\cmidrule(lr){3-5}
 &  & OS & Struct & Dist & & \\
\midrule
\multirow{3}{*}{OS vs. Struct} 
 & Relevance & 8 & 4 & -- & 0.388 & No \\
 & Fabrication & 5 & 7 & -- & 0.774 & No \\
 & Actionability & 7 & 5 & -- & 0.774 & No \\
\midrule
\multirow{3}{*}{OS vs. Distill} 
 & Relevance & 9 & -- & 3 & 0.146 & No \\
 & Fabrication & 6 & -- & 6 & 1.000 & No \\
 & Actionability & 9 & -- & 3 & 0.146 & No \\
\midrule
\multirow{1}{*}{Struct vs. Dist} 
 & Fabrication & -- & 5 & 7 & 0.774 & No \\
\bottomrule
\end{tabular}
\caption{Clinician preferences between  summaries. No significant differences were found across comparisons (p > 0.05). OS: One-step, Struct: Structured, Dist: Distilled.}
\label{tab:clinician_preferences}
\end{table}

\begin{table}[t!]
\centering
\begin{tabular}{lccc}
\toprule
\textbf{Metric} & \textbf{System} & \textbf{LLM rank} & \textbf{Clinician rank} \\
\midrule
\multirow{3}{*}{Relevance}
  & One-step   & 1 & 1 \\
  & Distilled  & 2 & 3 \\
  & Structured & 3 & 2 \\
\midrule
\multirow{3}{*}{Factual Fabrication}
  & Distilled  & 1 & 1 \\
  & One-step   & 2 & 3 \\
  & Structured & 3 & 2 \\
\midrule
\multirow{3}{*}{Actionability}
  & One-step   & 1 & 1 \\
  & Structured & 2 & 2 \\
  & Distilled  & 3 & 3 \\
\midrule
\multirow{3}{*}{Overall (avg rank)}
  & One-step   & 1.33 & 1.67 \\
  & Distilled  & 2.00 & 2.33 \\
  & Structured & 2.67 & 2.00 \\
\midrule
\multicolumn{3}{r}{\textbf{Spearman rank correlation ($\rho$)}} & 0.67 \\
\multicolumn{3}{r}{\textbf{p-value}} & 0.050 \\
\bottomrule
\end{tabular}
\caption{Comparison of LLM and clinician preferences by evaluation metric. Rankings are ordinal (1 = best). Spearman’s $\rho$ indicates a strong positive correlation between preferences.}
\label{tab:llm_clinician}
\end{table}

\subsection{Limitations}

In this study, we outlined and investigated the feasibility of the \texttt{DistillNote} framework for LLM clinical summary evaluation. Heart failure diagnosis was chosen as our functional task because it requires combining a plethora of clinical signals, making it an ideal test for information retention verification. Despite using a single dataset, MIMIC-IV, as an extensive and benchmark dataset, it provided the necessary scale for validating the framework at scale. Having demonstrated the methodological feasibility of \texttt{DistillNote} through heart failure prediction across three LLMs and three summary styles, the approach can be easily adapted to broader multi-condition, multi-dataset applications, requiring only a change in the diagnosis prompt and the prediction golden labels.

Furthermore, clinician evaluation was conducted on 18 clinical notes by two doctors. Although this covers a limited scope compared to the summary dataset's size, it offers empirical insights into how clinicians perceive AI-generated summaries. Obtaining detailed expert assessments of clinical summaries is still resource-intensive and methodologically challenging, especially at scale, further underscoring the motivation for better automatic functional evaluation.

\newpage
\onecolumn
\subsection{Clinical summary evaluation form guidelines}
\footnotesize
\begin{verbatim}
Purpose: This study aims to evaluate the quality of admission note summaries generated with
large language models (LLMs) using different approaches. As a clinical expert, your
assessment will help determine which approach produces the best summaries across three metrics.

Task: We kindly request your assessment of 18 cases. For each case, you will be presented with:
1. An original admission note
2. Two different summaries (labeled A and B)
Your task is to compare the two summaries and indicate which one you prefer based on the metrics.

Evaluation Criteria:
- Clinical relevance: Which summary better captures and preserves the medically important
information from the original note, maintaining appropriate focus on key clinical findings?
- Factual fabrication: Which summary contains fewer factual errors, made-up information, or
hallucinations compared to the source note? Consider whether all statements are directly
supported by the original note.
- Clinical actionability: Which summary would be more useful for clinical decision-making,
providing clearer information
for next steps, treatment planning, or handoffs at admission time?

Important Notes: 
- The summaries are presented in random order and are unlabeled as to their source. 
- There are no "right" answers - we are interested in your professional clinical judgment. 
- Both summaries may have strengths and weaknesses - please select the one you consider superior.
- Try your best to choose either A or B. However, if you find the two summaries to be equal in
quality based on a specific
metric, mention the "tie" in the comment section.
- Please complete all 18 cases.
\end{verbatim}

\newpage
\onecolumn
\subsection{LLM-as-judge prompt template: Relevance}
\label{relevance}
\footnotesize
\begin{verbatim}
You are an expert in evaluating medical summaries. Your task is to assign a score for the
following admission note summary based on the admission note, using the specified evaluation
criteria.

IMPORTANT INSTRUCTIONS:
- This is specifically evaluating an ADMISSION NOTE SUMMARY - a concise summary of a patient's
initial hospital admission record.
- Provide ONLY a single decimal number as your response, with NO explanation or additional text.
- The score MUST include a decimal point (e.g., 4.3, 3.7, 2.2). Do not use whole numbers.
- DO NOT add extra words or explanations after the score.

# INPUT (Original Admission Note)
<inputs>
[ADMISSION NOTE]
</inputs>

# OUTPUT (Admission Note Summary to be evaluated)
<output>
[SUMMARY TO BE EVALUATED]
</output>

# EVALUATION CRITERIA FOR ADMISSION NOTE SUMMARY
<evaluation_criteria>
Clinical Relevance (1.0-5.0): Measures how completely and accurately the admission note summary
captures key medical facts from the original admission note. This evaluation is specifically for
an admission note summary, which should capture the essential information documented during a
patient's initial hospital admission.
</evaluation_criteria>

EVALUATION GUIDELINES FOR ADMISSION NOTE SUMMARIES:
- Focus on whether all CLINICALLY IMPORTANT information from the note is included
- Consider whether the information is appropriately prioritized based on clinical significance
- Assess if the summary captures the essential patient history, chief complaint, presentation,
initial findings, and preliminary diagnoses
- Look for appropriate inclusion of vital signs, lab values, and test results documented at
admission that influence care
- Consider whether medication information and allergies with clinical impact are included
- Do NOT overly penalize for omitting minor details that don't affect initial clinical care
decisions

# SCORING RUBRIC
<scoring_rubric>
5.0: Perfect clinical relevance - captures all critical information with perfect prioritization
4.0-4.9: Very good clinical relevance - captures most critical information with good
prioritization
3.0-3.9: Adequate clinical relevance - captures basic information with acceptable
prioritization
2.0-2.9: Poor clinical relevance - misses important elements or prioritizes less relevant
information
1.0-1.9: Very poor clinical relevance - misses most critical information
</scoring_rubric>

# CALIBRATED EXAMPLES OF ADMISSION NOTE SUMMARIES

## EXAMPLE 1: Score 5.0 (High Quality)
/synthetic_admission_note>
Example 1 (high)
</synthetic_admission_note>

<admission_note_summary>
Example 1 (high)
</admission_note_summary>

<rationale>
Example 1 (high)
</rationale>

## EXAMPLE 2: Score 3.5 (Medium Quality)
/synthetic_admission_note>
Example 2 (medium)
</synthetic_admission_note>

<admission_note_summary>
Example 2 (medium)
</admission_note_summary>

<rationale>
Example 2 (medium)
</rationale>

## EXAMPLE 3: Score 1.4 (Low Quality)
/synthetic_admission_note>
Example 3 (low)
</synthetic_admission_note>

<admission_note_summary>
Example 3 (low)
</admission_note_summary>

<rationale>
Example 3 (low)
</rationale>

# YOUR TURN - PROVIDE ONLY A SINGLE DECIMAL SCORE BELOW FOR THIS ADMISSION NOTE SUMMARY

Clinical Relevance:
\end{verbatim}
\onecolumn
\subsection{LLM-as-judge prompt template: Factual Fabrication}
\label{fabrication}
\footnotesize
\begin{verbatim}
You are an expert in evaluating medical summaries. Your task is to assign a score for the
following admission note summary based on the admission note, using the specified evaluation
criteria.

IMPORTANT INSTRUCTIONS:
- This is specifically evaluating an ADMISSION NOTE SUMMARY - a concise summary of a patient's
initial hospital admission record.
- Provide ONLY a single decimal number as your response, with NO explanation or additional text.
- The score MUST include a decimal point (e.g., 4.3, 3.7, 2.2). Do not use whole numbers.
- DO NOT add extra words or explanations after the score.

# INPUT (Original Admission Note)
<inputs>
[ADMISSION NOTE]
</inputs>

# OUTPUT (Admission Note Summary to be evaluated)
<output>
[SUMMARY TO BE EVALUATED]
</output>

# EVALUATION CRITERIA FOR ADMISSION NOTE SUMMARY
<evaluation_criteria>
Factual Fabrication (1.0-5.0): Measures ONLY whether the note summary introduces facts that are
completely fabricated or invented and cannot be traced to or reasonably inferred from the
original admission note. This evaluation is specifically for an admission note summary, which
should capture the essential information documented during a 
patient's initial hospital admission.
</evaluation_criteria>

EVALUATION GUIDELINES FOR ADMISSION NOTE SUMMARIES:
- Focus ONLY on identifying information that is purely invented with no basis in the note
- Do NOT penalize for:
  * Reasonable clinical interpretations or conclusions based on information in the note
  * Organization or categorization of information present in the note
  * General demographic descriptions
  * Implied severity or acuity that matches clinical findings documented in the note
  * Standard medical terminology used to describe conditions mentioned in the note
- DO penalize for:
  * Adding medical conditions not mentioned in the note
  * Inventing specific test results, vital signs, or measurements not in the note
  * Creating patient history elements with no basis in the note
  * Stating specific treatments were given when not mentioned in the note
  * Making definitive statements about prognosis or outcomes not supported by the note

# SCORING RUBRIC
<scoring_rubric>
5.0: No fabrication - every statement is directly supported by or can be reasonably inferred
from the note.
4.0-4.9: Minimal fabrication - contains only 1-2 minor details that might be slight
overextensions but do not contradict the note.
3.0-3.9: Some fabrication - contains a few statements that have no basis in the note but do not
significantly alter the clinical picture.
2.0-2.9: Substantial fabrication - contains multiple statements that are entirely invented with
no support in the note.
1.0-1.9: Pervasive fabrication - contains critical invented information that fundamentally
misrepresents the patient's condition as documented in the note.
</scoring_rubric>

# CALIBRATED EXAMPLES OF ADMISSION NOTE SUMMARIES

## EXAMPLE 1: Score 5.0 (High Quality)
/synthetic_admission_note>
Example 1 (high)
</synthetic_admission_note>

<admission_note_summary>
Example 1 (high)
</admission_note_summary>

<rationale>
Example 1 (high)
</rationale>

## EXAMPLE 2: Score 3.8 (Medium Quality)
/synthetic_admission_note>
Example 2 (medium)
</synthetic_admission_note>

<admission_note_summary>
Example 2 (medium)
</admission_note_summary>

<rationale>
Example 2 (medium)
</rationale>

## EXAMPLE 3: Score 1.2 (Low Quality)
/synthetic_admission_note>
Example 3 (low)
</synthetic_admission_note>

<admission_note_summary>
Example 3 (low)
</admission_note_summary>

<rationale>
Example 3 (low)
</rationale>

# YOUR TURN - PROVIDE ONLY A SINGLE DECIMAL SCORE BELOW FOR THIS ADMISSION NOTE SUMMARY

Factual Fabrication:
\end{verbatim}

\newpage
\onecolumn
\subsection{LLM-as-judge prompt template: Clinical Actionability}
\label{actionability}
\footnotesize
\begin{verbatim}
You are an expert in evaluating medical summaries. Your task is to assign a score for the
following admission note summary based on the admission note, using the specified evaluation
criteria.

IMPORTANT INSTRUCTIONS:
- This is specifically evaluating an ADMISSION NOTE SUMMARY - a concise summary of a patient's
initial hospital admission record.
- Provide ONLY a single decimal number as your response, with NO explanation or additional text.
- The score MUST include a decimal point (e.g., 4.3, 3.7, 2.2). Do not use whole numbers.
- DO NOT add extra words or explanations after the score.

# INPUT (Original Admission Note)
<inputs>
[ADMISSION NOTE]
</inputs>

# OUTPUT (Admission Note Summary to be evaluated)
<output>
[SUMMARY TO BE EVALUATED]
</output>

# EVALUATION CRITERIA FOR ADMISSION NOTE SUMMARY
<evaluation_criteria>
Clinical Actionability (1.0-5.0): Measures how clearly, concisely, and effectively the
admission note summary presents urgent or decision-critical information to support clinical
decision-making at the time of hospital admission. This evaluation is specifically for an
admission note summary, which should capture the essential information documented
during a patient's initial hospital admission.
</evaluation_criteria>

EVALUATION GUIDELINES FOR ADMISSION NOTE SUMMARIES:
- Focus on how well the summary facilitates immediate clinical decisions at the time of
admission
- Consider if critical information from the note is highlighted prominently
- Assess whether the organization helps prioritize initial clinical concerns
- Look for clear presentation of abnormal findings
- Consider if medication information, allergies, and contraindications are presented
- Evaluate whether the format supports rapid understanding of the patient's status
- Consider if next steps or needed interventions are clearly implied by the information
presented

# SCORING RUBRIC
<scoring_rubric>
5.0: Perfect clinical actionability - optimally presents all decision-critical information
4.0-4.9: Very good clinical actionability - presents most decision-critical information
with good organization
3.0-3.9: Adequate clinical actionability - presents important information but with suboptimal
organization
2.0-2.9: Poor clinical actionability - presents some information but with poor prioritization
1.0-1.9: Very poor clinical actionability - insufficient information for clinical
decision-making
</scoring_rubric>

# CALIBRATED EXAMPLES OF ADMISSION NOTE SUMMARIES

## EXAMPLE 1: Score 5.0 (High Quality)
/synthetic_admission_note>
Example 1 (high)
</synthetic_admission_note>

<admission_note_summary>
Example 1 (high)
</admission_note_summary>

<rationale>
Example 1 (high)
</rationale>

## EXAMPLE 2: Score 3.4 (Medium Quality)
/synthetic_admission_note>
Example 2 (medium)
</synthetic_admission_note>

<admission_note_summary>
Example 2 (medium)
</admission_note_summary>

<rationale>
Example 2 (medium)
</rationale>

## EXAMPLE 3: Score 1.3 (Low Quality)
/synthetic_admission_note>
Example 3 (low)
</synthetic_admission_note>

<admission_note_summary>
Example 3 (low)
</admission_note_summary>

<rationale>
Example 3 (low)
</rationale>

# YOUR TURN - PROVIDE ONLY A SINGLE DECIMAL SCORE BELOW FOR THIS ADMISSION NOTE SUMMARY

Clinical Actionability:


\end{verbatim}

\end{document}